\documentclass[accepted]{mystyle} % after acceptance, for a revised version; 
\usepackage[american]{babel}
\usepackage{natbib} % has a nice set of citation styles and commands
\usepackage{mathtools} % amsmath with fixes and additions
\usepackage{booktabs} % commands to create good-looking tables
\usepackage{tikz} % nice language for creating drawings and diagrams

\usepackage{algorithm}
\usepackage{algpseudocode}

\usepackage{tabularx}
\usepackage{multirow}

\usepackage{amsfonts}       % blackboard math symbols
\usepackage{nicefrac}       % compact symbols for 1/2, etc.
\usepackage{microtype}      % microtypography
\usepackage{xcolor}         % colors
\usepackage{amsmath}
\usepackage{amssymb}
\usepackage{physics}
\usepackage{graphicx}
\usepackage{bbm}
\usepackage{amsthm}

\DeclareMathOperator{\bbv}{\boldsymbol{v}}
\DeclareMathOperator{\bx}{\boldsymbol{x}}
\DeclareMathOperator{\by}{\boldsymbol{y}}

\DeclareMathOperator{\btheta}{\boldsymbol{\theta}}

\DeclareMathOperator{\dbx}{\textrm{d}\boldsymbol{x}}

\DeclareMathOperator{\bG}{\boldsymbol{G}}
\DeclareMathOperator{\bI}{\boldsymbol{I}}

\DeclareMathOperator{\bX}{\boldsymbol{X}}

\DeclareMathOperator{\bLambda}{\boldsymbol{\Lambda}}

\DeclareMathOperator{\cH}{\mathcal{H}}
\DeclareMathOperator{\cN}{\mathcal{N}}

\DeclareMathOperator{\Ex}{\mathbb{E}}
\DeclareMathOperator{\man}{\mathcal{M}}
\DeclareMathOperator{\dS}{\mathbb{S}}
\DeclareMathOperator{\N}{\mathbb{N}}
\DeclareMathOperator{\R}{\mathbb{R}}
\DeclareMathOperator{\diag}{\mathrm{diag}}

\newcommand{\inp}[2]{\left\langle #1, #2 \right\rangle}

\newtheorem{proposition}{Proposition}
\newtheorem{observation}{Observation}

\newcommand{\ournamefull}{Metric-agnostic Geodesic Slice Sampler}
\newcommand{\ourmethod}{MAGSS}
\newcommand{\metaourmethod}{Meta-MAGSS}

\title{
Geodesic Slice Sampler for Multimodal Distributions with Strong Curvature}

\author[1]{\href{mailto:<bernardo.williamsmoreno@helsinki.fi>?Subject=Geodesic Slice Sampler for Multimodal Distributions with Strong Curvature}{Bernardo Williams}{}}
\author[1]{Hanlin Yu}
\author[1]{Hoang Phuc Hau Luu}
\author[2]{Georgios Arvanitidis}
\author[1]{Arto Klami}
\affil[1]{%    
    Department of Computer Science, University of Helsinki, Finland    
}
\affil[2]{%
    Cognitive Systems, DTU Compute, Technical University of Denmark
}
  
\begin{document}

% Do not add contents to table of contents
\addtocontents{toc}{\protect\setcounter{tocdepth}{0}}

    \maketitle
    
\begin{abstract}    
Traditional Markov Chain Monte Carlo sampling methods often struggle with sharp curvatures, intricate geometries, and multimodal distributions. Slice sampling can resolve local exploration inefficiency issues, and Riemannian geometries help with sharp curvatures. Recent extensions enable slice sampling on Riemannian manifolds, but they are restricted to cases where geodesics are available in a closed form. We propose a method that generalizes Hit-and-Run slice sampling to more general geometries tailored to the target distribution, by approximating geodesics as solutions to differential equations. Our approach enables the exploration of the regions with strong curvature and rapid transitions between modes in multimodal distributions. We demonstrate the advantages of the approach over challenging sampling problems.
\end{abstract}

\section{Introduction}\label{sec:intro}

Sampling from a differentiable unnormalized log-density defined on a Euclidean space is a core problem in machine learning and statistics. While gradient-based Markov Chain Monte Carlo (MCMC) methods have proven effective in many scenarios, they often face significant challenges when the target distribution exhibits complex geometry (sharp curvature) or multimodal behavior. The two core challenges are largely addressed with complementary techniques, with little work on algorithms that excel for targets that are \emph{both} multimodal and complex in shape.

Complex shapes and sharp curvatures are often addressed by using a suitably chosen Riemannian geometry within the sampling algorithms \citep{Girolami2011}. Instead of operating in a Euclidean space and metric, the samplers carry out the necessary operations using a metric that adapts to the curvature of the parameter space. In practice, the methods follow flows induced by the metric, in most cases by numerical integration, and consequently the methods are sometimes called \emph{geodesic} methods as in our title. Various practical metrics and sampling algorithms have been shown to improve the sampling of targets with strong curvature \citep{Girolami2011,Byrne2013,Lan2015,Hartmann2022,Hartmann2023,Williams2024}, albeit always with increased computational cost.

Multimodality, in turn, is most commonly addressed by tempering or diffusion techniques \citep{Earl2005,Chen2024}. These methods use a tempered (smoothed) version of the target to improve exploration over multiple modes, intuitively changing the problem itself so that the modes are connected with areas of sufficient probability. At a high degree of tempering these methods can efficiently explore the different modes, but low tempering is needed for accurate sampling within the modes, necessitating adaptive or parallel sampling with different degrees of tempering.
The efficiency of parallel tempering depends on the swap acceptance rate between adjacent temperatures, which can decrease in high dimensions if the temperature schedule is not well-tuned \citep{Woodard2009}.  
Diffusion-based approaches, in turn,  require careful choice of the noise schedule to balance exploration and accuracy \citep{Song2019, Chen2024}. Unlike tempering, diffusion methods can achieve smooth transitions between modes without explicitly maintaining a set of parallel chains, but the acceptance rate of noisy samples can be low \citep{Chen2024}.

Even though the two approaches are efficient in addressing the two challenges separately, there is very little work on samplers designed for the general setup where both difficulties may arise simultaneously. One could consider e.g. parallel tempering in a Riemannian metric --- see \citet{Byrne2013} for a rare example in this intersection --- but ideally we would like to address both aspects using a common mechanism. This work explores one such approach, developing a Riemannian sampler capable of efficiently exploring multiple modes, without any tempering for the target distribution. Instead, we seek to improve mode exploration by changing  the metric, in the spirit of the early work by \citet{Lan2014} that developed a specific metric solely for this purpose. Their metric, however, requires explicit identification and tracking of the modes and is more like a conceptual demonstration, and we are not aware of any other works aiming for efficient multimodal samplers solely by the change of the metric.

% We are motivated by the idealized slice sampler, with computable level sets.
% The performance of such an idealized sampler is theoretically independent of the sampling problem, as pointed out by \citet{Durmus2023}:
% \textsl{“This means that the performance of the idealized slice sampler is ignorant of the introduction of, e.g., multimodality, local modes, or anisotropy as long as the volume of the level sets is not modified.”}
% If we change the geometry of the problem such that level sets are easier to compute, then the slice sampler should benefit.
We are motivated by the idealized slice sampler with computable level sets. As noted by \citet{Durmus2023}:
\textsl{“This means that the performance of the idealized slice sampler is ignorant of the introduction of, e.g., multimodality, local modes, or anisotropy as long as the volume of the level sets is not modified.”}
This insight suggests that by modifying the geometry of the problem to produce simpler or more tractable level sets, the slice sampler can effectively handle multimodal distributions.
From a practical perspective, we build on the (Euclidean) Hit-and-Run slice sampler by \citet{Belisle1993}, which at each iteration selects a random direction and then samples from the resulting one-dimensional distribution formed by the intersection of the line and the slice. 
In effect, it transforms multi-dimensional sampling into sequential one-dimensional sampling tasks, but the overall sampler may be inefficient. Especially in higher dimensions, the intersection with the slice can be small for almost all directions \cite{Murray2010}.

Both \citet{Habeck2023} and \citet{Durmus2023} recently considered generalizations of the Hit-and-Run sampler for Riemannian manifolds, replacing the lines with geodesics. We build on the general algorithmic framework introduced by \citet{Durmus2023} and adapt it to the task of sampling from a distribution with a complex geometry. Specifically, we begin by embedding the (Euclidean) sampling space into a higher-dimensional space that incorporates the target distribution’s geometric information, such as Fisher information or Monge embedding \citep{Hartmann2022}. This transforms the problem into sampling from a particular Riemannian manifold where the target distribution corresponds to the Hausdorff density (see Section \ref{sect:approx_geo}). 
Note that even though we leverage components proposed by \citet{Durmus2023}, our task is fundamentally more difficult. Their starting point was sampling of a density on a known manifold (e.g., a sphere) where the geodesics are  exactly known, whereas 
the complexity of our embedding manifold requires us to approximate the geodesics using numerical integrators.

\begin{figure}[t]
    \centering
    \includegraphics[width=\linewidth]{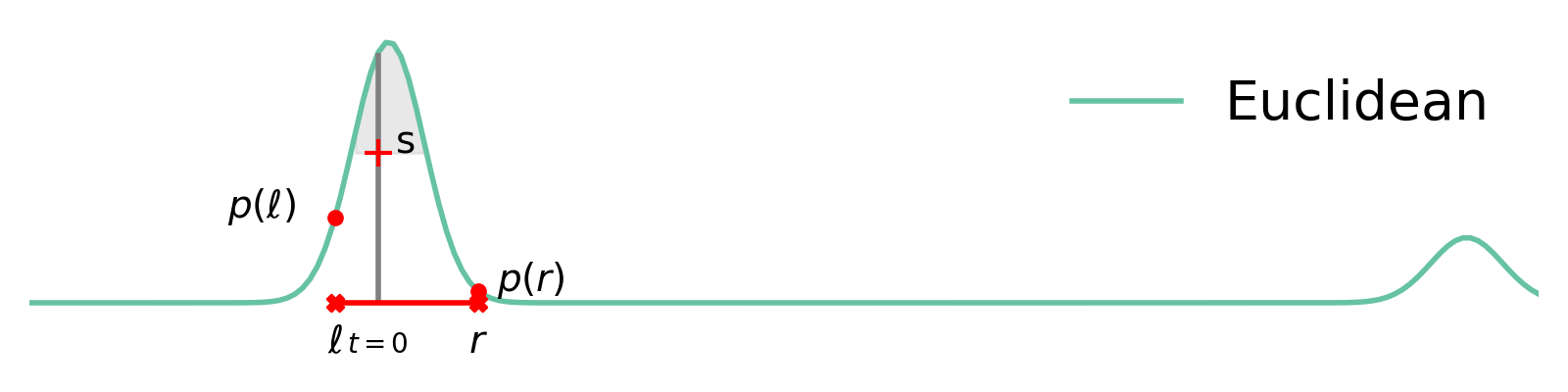}
    \includegraphics[width=\linewidth]{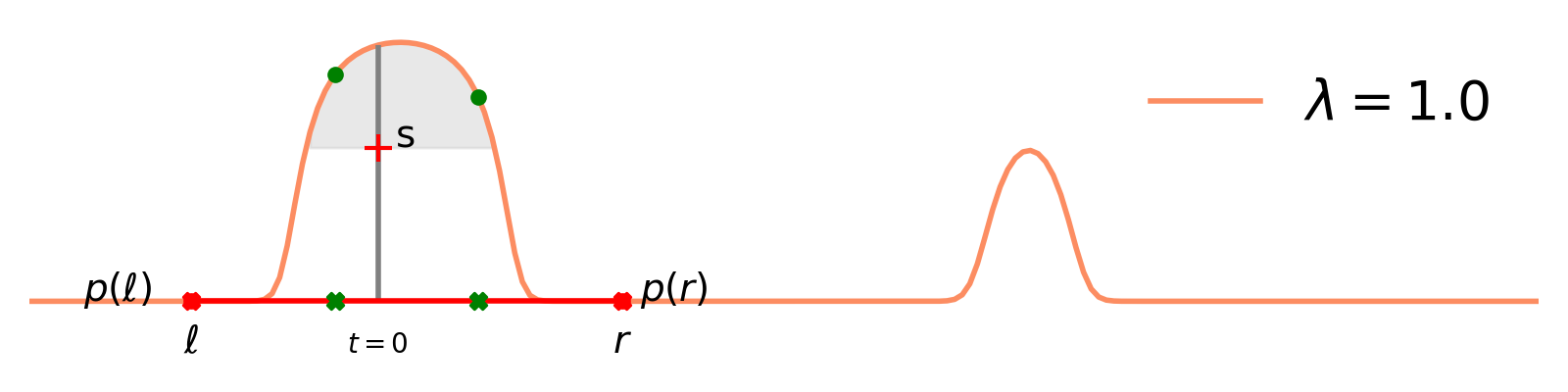}
    \includegraphics[width=\linewidth]{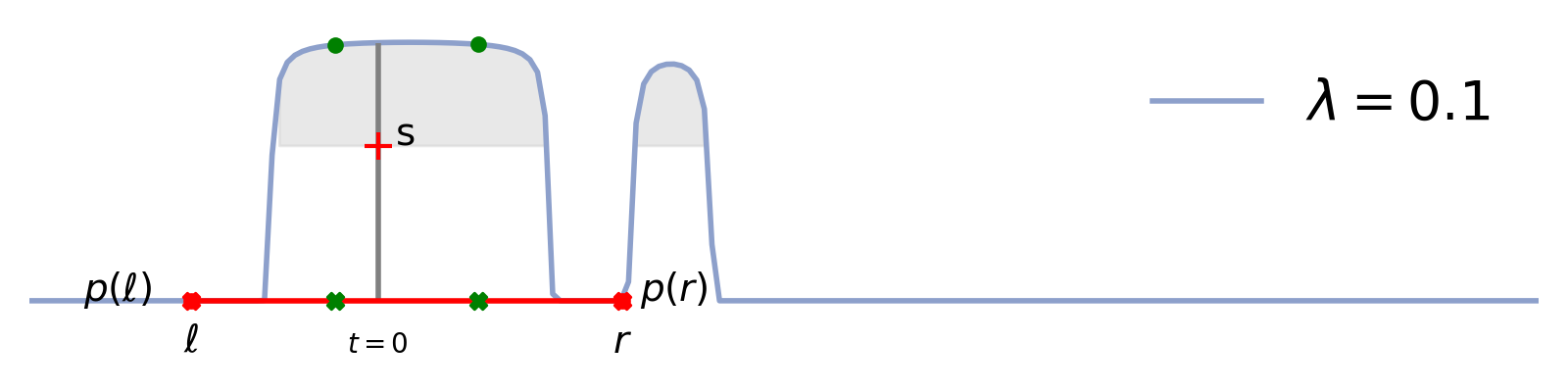}
    \includegraphics[width=\linewidth]{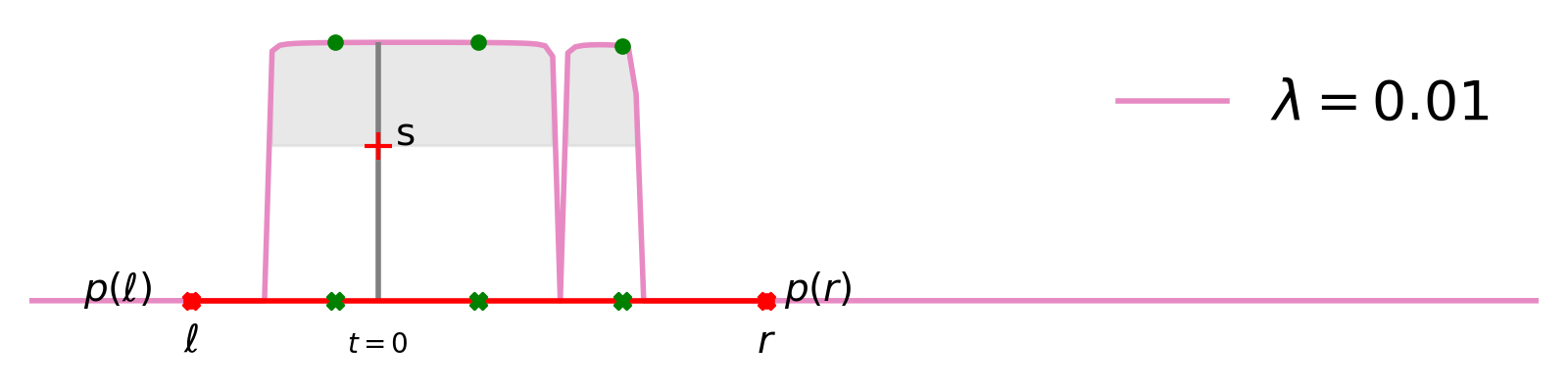}
    \caption{Illustration of the step-out procedure in \ournamefull.
    The lines drawn with different colors represent the Hausdorff density $p(t) := p_{\cH}(\hat{\gamma}_{(\bx, \bbv)}(t))$ (Eq.~\eqref{eq:hausdensity}) considering the Inverse Generative metric for different values of $\lambda$ (Eq.~\eqref{eq:gen}).
    The step-out procedure chooses  $s \sim \mathrm{Unif}(0,p(0))$ and sets randomly an interval of length $r-\ell$ at $t=0$ with left and right points $\ell$ and $r$. While $p(r) > p(s)$ it expands the right side of the interval as $r = r + w$ and for the left side while $p(l) > p(s)$ it does $\ell = \ell - w$. This expands the length of the initial interval.     
    As $\lambda \rightarrow 0$ the space shrinks due the properties of the metric, making it easier for the step-out procedure to jump to the distant mode.        
    }
    \label{fig:fig1}
\end{figure}

In this work, we propose a geodesic slice sampler applicable for arbitrary Riemannian metrics, and discuss the choice of the metric. In particular, we introduce two new computationally efficient metrics. Both metrics improve sampling over multimodal targets by, in a sense, pulling the modes closer to each other; see Figure~\ref{fig:fig1} illustrating this effect within the slice sampler, as a function of a parameter $\lambda$ controlling how much the metric warps the space.
% see Figure~\ref{fig:fig1} and Figure~\ref{fig:fig2} illustrating this effect. The former within the slice sampler in a single dimensional space, as a function of a parameter $\lambda$ controlling how much the metric warps the space. And the latter in a two dimensional space.
In addition, we introduce a meta-sampler similar to \citet{Tjelmeland2001} that combines the proposed method with a separate sampler for improved exploration of local modes.

We empirically demonstrate  improved sampling over Euclidean methods for complex targets, and highlight improved mixing over multiple modes in high dimensional-cases when compared against  parallel tempering \citep{Swendsen1986,Latuszynski2025} and the diffusive Gibbs sampler by \citet{Chen2024} designed for addressing multimodality. Similar to previous Riemannian methods, the algorithm shows good exploration and mixing, but has slower iterations because of the numerical computation of the geodesics. 

\section{Background: Slice sampling}

The classic work of \citet{Neal2003} introduces slice sampling as a method for generating samples by uniformly sampling from the $\R^{D+1}$ manifold defined by the graph of the probability density.
Let $p(\bx)$ be an unnormalized continuous target density that satisfies $\int p(\bx) \dbx < \infty$. Suppose that direct sampling from $p(\bx)$ is not feasible.
We consider densities where $\bx \in \R^D$ with respect to the Lebesgue measure.

Idealized slice sampling defines a uniform distribution over the volume under the graph of $p(\bx)$ and generates samples through the following two steps: 
\begin{enumerate}
    \item Sample $s \sim \mathrm{Unif}(0, p(\bx))$.
    \item Sample $\bx \sim \mathrm{Unif}( L(s) )$.
\end{enumerate}
where the slice is given by $L(s) := \{ \bx \mid p(\bx) > s \}$. For special cases, such as log-concave or rotationally invariant densities, the slice sampler has theoretical performance guarantees \citep{Natarovskii2021}. However, for more complex distributions, drawing uniform samples from $L(s)$ is often impractical \citep{Rudolf2018}.

To address this, the step-out and shrinkage procedures are used. Below, we provide an informal explanation of these procedures. The full algorithm is detailed in the Appendix (Algorithms \ref{alg:stepout} and \ref{alg:shrink}).
Both procedures were first introduced by \citet{Neal2003}, but we adopt an equally valid modified version of the shrinkage step as proposed by \citet{Durmus2023}.
%Both versions are equally valid. 
For a moment, assume a univariate density $p(x)$ and a current position $x \in \R$. The procedures are as follows:

\paragraph{The Step-Out Procedure}  
The step-out procedure, illustrated in Figure \ref{fig:fig1}, takes two parameters: the width $w \in \R$ and maximum steps $m \in \N$. Given the slice $L(s)$, the goal is to expand an interval around the current point $x$.  Consider the auxiliary function $\gamma_{x}(t)=x + t$.

The initial left $\ell$ and right $r$ points are set at a random distance $w$ apart. This is done by sampling $u \sim \mathrm{Unif}(0, w)$ and setting $\ell = -u$ and $r = \ell + w$. To ensure that at most $m+1$ expansion steps are performed (combined for both directions), a random integer $\iota \sim \mathrm{Unif}(\{1, \dots, m\})$ is sampled. The right limit is expanded up to $\iota$ times, and the left limit up to $m+1-\iota$ times.  

The expansion proceeds as follows:
The right limit $r$ is expanded by adding $w$ until $p(\gamma_x(r + w)) < s$, meaning $\gamma_x(r + w) \notin L(s)$. The left limit $\ell$ is expanded by subtracting $w$ until $p(\gamma_x(\ell - w)) < s$, meaning $\gamma_x(\ell - w) \notin L(s)$.  
The procedure returns the updated interval $(\ell, r)$. We denote it by $\text{Step-out}_{w,m}(s, \gamma_x)$.

\paragraph{The Shrinkage Procedure}  
The shrinkage procedure selects a sample from the interval $(\ell, r)$ by gradually reducing its size until a point is found within $L(s) \cap (\ell, r)$.  

The interval $J = (\ell, r)$ is treated as a circular domain, meaning that if we move past $r$, we continue from $\ell$. The procedure starts by sampling two points $y$ and $z$ uniformly within $(\ell, r)$. If neither $\gamma_x(y)$ or $\gamma_x(z)$ fall inside $L(s)$, the interval is shrunk as follows:  
\begin{itemize}
    \item Form the interval $(y\wedge z, y\vee z )$. Update the circular region by
    \begin{equation*}
        J =
        \begin{cases} 
        J\cap(y \wedge z, y \vee z), & \text{if } 0 \in J, \\
        J \setminus (y \wedge z, y \vee z), & \text{if } 0 \notin J.
        \end{cases}        
    \end{equation*}    
    \item Set $y=z$ and  update $z\sim \mathrm{Unif}(J)$.  
    \item This process repeats, each time reducing the size of the interval, until $\gamma_x(z) \in L(s)$.
\end{itemize}
We denote the procedure $\text{Shrink}_{l,r}(s, \gamma_x)$.
One complete step of the slice sampler is:
\begin{enumerate}
    \item Sample $s \sim \mathrm{Unif}(0,p(x))$
    \item Obtain $\ell,r = \text{Step-out}_{w,m}(s,\gamma_x)$ 
    \item Sample $t^* = \text{Shrink}_{\ell,r}(s, \gamma_x)$.
    \item Set $x = \gamma_x(t^*)$.
\end{enumerate}

\paragraph{Hit-and-Run} 
One way to extend slice sampling to multivariate distributions is to combine it with  Hit-and-Run sampling,
 presented here following \citet{Belisle1993}. Let $ \dS^{D-1}(\bx) = \{\bbv \in \R^D : \norm{\bbv}^2 =1\}$, and  $\bbv \sim \mathrm{Unif}( \dS^{D-1}(\bx) )$. An iteration of the whole sampler is: 
\begin{itemize}
    \item Obtain $\bbv\sim \mathrm{Unif}( \dS^{D-1}(\bx) )$.
    \item Obtain a sample from the density evaluated on the straight line (Euclidean geodesic) 
    \begin{equation*}
     t \mapsto p(\bx + t \bbv)/ \int p(\bx + t \bbv) \dd t.
    \end{equation*}

\end{itemize}
When directly sampling a value $t$ according the density along a straight line $t \mapsto  p(\bx + t \bbv)/ \int p(\bx + t \bbv) \dd t$ is not feasible, we can use slice sampling on $t \mapsto  p(\bx + t \bbv)$, since it is an unnormalized univariate distribution. Define $\gamma_{(x,v)}(t) = \bx + t \bbv$.
The step-out procedure outputs $\ell,r = \text{Step-out}_{w,m}(s, \gamma_{(x,v)})$ and
the shrinkage procedure will return $t^* = \text{Shrink}_{\ell,r}(s, \gamma_{(x,v)})$. The new sample is $\bx =  \gamma_{(x,v)}(t^*)$.
This is called  Hit-and-Run slice sampling or hybrid slice sampling \citep{Latuszynski2014}. This method extends slice sampling to probability distributions defined over $\R^D$.

\section{Method}
%\ournamefull}
\label{sect:approx_geo}
Our main contribution is a geodesic slice sampler that can accommodate arbitrary metrics. It extends the Hit-and-Run slice sampler described above for non-Euclidean geometries, similar to the recent works of \citet{Durmus2023} and \citet{Habeck2023}, but instead of leveraging closed-form analytic geodesics of predefined manifolds we induce metrics using characteristics of the target density itself to guide the sampling. Now the geodesics need to be approximated by numerical integrators.
This section explains the sampler and a meta-sampler that combines the core method with separate local sampler for improved efficiency for general metrics, always using $\bG(\bx)$ to denote the metric tensor. We will discuss specific metrics in Section~\ref{sec:metrics}.

Straight-line Hit-and-Run sampling can be inefficient because proposals often move away from high-probability regions \citep{Murray2010}. To resolve this, we perform slice sampling along geodesic curves that can accommodate the geometry of the target distribution. This improves efficiency when the target distribution is highly curved or multimodal; see Section~\ref{sec:metrics_global}. 
We are interested in sampling problems defined in $\mathbb{R}^D$ but allow using different plug-in metrics (preferably using the target density information) to enhance exploration. 
% This can be cast as sampling from a distribution defined on a Riemannian manifold where the algorithm in \citet{Durmus2023} can be employed. 

The general problem can be cast as sampling from a distribution defined on a Riemannian manifold where we adapt the algorithm of \citet{Durmus2023} under general  metrics.
Alternatively, it can be seen as Hit-and-Run where straight lines are replaced by curves that better wrap around the level sets of the target density (given the metrics are good enough). Because the metric is general, closed-form geodesics are unavailable, so we must compute them with numerical integrators.
To correctly sample along geodesics, we need three key components:  Adjusting for the correct density on the manifold,  properly sampling directions using the Riemannian metric, and solving the geodesic equations. 

\paragraph{Hausdorff Density: } 
To ensure we sample from the correct distribution on the manifold with metric $\bG(\bx)$, we must account for the change in measure from the Euclidean space to the manifold. The correct density is the Hausdorff density
\begin{equation} 
  p_{\cH}(\bx) = \frac{p(\bx)}{\sqrt{\det \bG(\bx)}}. \label{eq:hausdensity}
\end{equation}
The denominator adjusts for local volume distortion introduced by the metric $\bG(\bx)$, ensuring that the volume over the manifold is preserved and hence maintaining proper sampling behavior. See Appendix \ref{app:hausdorff} for further details.

\paragraph{Sampling from the Riemannian Unit Ball: } 
%Here we explain the mechanism found in \citep{Durmus2023}.
Instead of sampling a random direction in Euclidean space, we must now sample from the unit geodesic ball under the Riemannian metric, where we can directly use the method proposed by \citet{Durmus2023}. Given a position $\bx$, a velocity $\bbv$ is sampled as follows: First draw $\bbv \sim \mathcal{N}(\mathbf{0}, \bG^{-1}(\bx))$, and then normalize it to obtain a unit-length vector in the Riemannian metric with:
\begin{equation*}
\bbv \leftarrow \frac{\bbv}{\|\bbv\|_g}, \quad \text{where} \quad \|\bbv\|_g = \sqrt{\bbv^\top \bG(\bx) \bbv}.    
\end{equation*}
This ensures that the direction is uniformly distributed on the unit sphere under the metric $\bG(\bx)$. See Appendix \ref{app:agss} for additional implementation details.

\paragraph{Approximating Geodesic Curves}
Given a sampled velocity $\bbv$, we need to follow the geodesic curve starting at $\bx$ in direction $\bbv$. In general, the geodesic equation
\begin{align}
    \dot \bx_k &= \bbv_k, \nonumber \\
    \dot \bbv_k &= - \norm{\bbv}^2_{\Gamma^k}, \quad \mathrm{for}\ k = 1, \ldots, D. \label{eq:geoeqs}
\end{align}
where $\Gamma^k_{ij} = \tfrac{1}{2}g^{km} ( \partial_i g_{m j} +  \partial_j g_{i m} - \partial_m g_{i j})$,
does not have a closed-form solution for arbitrary $\bG(\bx)$. See more detail in appendix \ref{app:geoeq}.
Instead, we numerically approximate the exponential map $\gamma_{(\bx, \bbv)}(t)$ by solving these differential equations with an ordinary differential equation (ODE) solver, denoted as $\hat{\gamma}_{(\bx, \bbv)}(t)$. 
The choice of the metric determines the shape of geodesic trajectories, allowing the sampler to adapt to different target distributions; see Section~\ref{sec:metrics}.

Algorithm \ref{alg:agss} explains the full \ournamefull \ (\ourmethod). After sampling a velocity $\bbv$ from the unit Riemannian sphere, slice sampling is performed on the Hausdorff density evaluated along the numerical solution of the geodesic trajectory. The step-out and shrinkage procedures then determine the final sample.
\begin{algorithm}[H] 
    \caption{\ournamefull}
    \label{alg:agss}
    \textbf{Input:} Initial position $\bx^{[0]}$, metric tensor $\bG(\bx)$, and parameters $m\in \mathbb{N}$, $w\geq 0$.\\
    \textbf{Output:} $N$ samples $\bx^{[n]}$.
    \begin{algorithmic}[1]                        
        \For{$n \leftarrow 0, \dots, N-1$}
            \State Sample $s \sim \mathrm{Unif}(0, p_{\cH}(\bx^{[n]}))$
            \State Sample velocity $\bbv^{[n]} \sim \textrm{Unif}(\dS_g^{D-1}(\bx^{[n]}))$      
            \State Compute  $(\ell, r) = \text{Step-out}_{w,m}(s, \hat{\gamma}_{(\bx^{[n]}, \bbv^{[n]})})$ 
            \State Sample time $t^* = \text{Shrink}_{\ell,r}(s, \hat{\gamma}_{(\bx^{[n]}, \bbv^{[n]})} )$
            \State $\bx^{[n+1]}= \hat{\gamma}_{(\bx^{[n]},\bbv^{[n]})}(t^*)$
        \EndFor
    \end{algorithmic} 
\end{algorithm}

\subsection{Meta Sampler and Multimodality}

The sampler as described above is valid as such, but we also introduce a simple extension that can further improve sampling for multimodal targets with complex local structure.

Following \citet{Tjelmeland2001,Latuszynski2025}, we create a \emph{meta-sampler} that alternates between using \ourmethod\ for global moves and an arbitrary local MCMC for sampling within each mode.
To generate one sample, we first run $K$-steps of \ourmethod\ followed by $L$-steps of any local MCMC sampler. 
We refer to this combined strategy as \metaourmethod, detailed in Algorithm~\ref{alg:metaagss} (in Appendix). 
The main motivation for this hybrid strategy is to leverage gradient-based algorithms for fast exploration of the mode, to utilize their efficient mixing and fast per-iteration computation when they are sufficiently good for the local target. We could in principle use any sampler for the local part, including Riemannian samplers, but we in practice use standard Euclidean Metropolis-adjusted Langevin Algorithms (MALA) \citep{Roberts1996} in our experiments.

%\subsection{Detailed Balance}

% The detailed balance follows from  \citet{Durmus2023}, but additional care is needed because the geodesics need to be computed numerically. We state the key theorem below, with the proof and additional discussion in Appendix~\ref{app:proof}.

% %However, unlike \citet{Durmus2023}, which focuses on the case where the geodesics are given in closed-form, in our case the geodesics have to be obtained through numerical integrations.
% \mainthm \label{thm:detailed_balance}
% %In appendix \ref{app:proof} we discuss the validity of our method.

\section{Metrics} \label{sec:metrics}

The sampler is general, applicable for an arbitrary metric and only requiring $\bG(\bx)$ to be positive definite and vary continuously. By selecting an appropriate metric we can influence how the geodesics explore the space, controlling the overall sampling behavior. There is no single metric that is optimal for all targets, and the metrics proposed in the literature are motivated by complementary argumentation, with notable emphasis in computational efficiency.

Next we discuss the metric choice.
The literature has exclusively focused on metrics that improve local exploration for complex target distributions, with several practical solutions that we re-cap in Section~\ref{sec:metrics_local}. We then turn our attention on how to improve exploration of multiple modes, presenting novel metrics specifically designed for this in Section~\ref{sec:metrics_global}.

\subsection{For adapting to local curvature}
\label{sec:metrics_local}

\paragraph{The Fisher metric}
The Fisher Information Metric (FIM) is defined as the covariance of the score function, and was predominantly used in the early Riemannian methods \citep{Girolami2011} due to its close connection to estimation theory. A general form of the metric is:
\begin{equation*}
\bG_F(\bx)  = \Ex_{\by|\bx}\left[ \nabla_{\bx} \log p(\by | \bx)  \nabla_{\bx} \log p(\by | \bx)  ^\top \right],
\end{equation*}
but the specific form depends on the underlying problem, due to integration over the conditional density. Furthermore, it requires direct matrix inversion for computing $\bG_F^{-1}(\bx)$ that is required during geodesic computations (Eq.~\ref{eq:geoeqs}), with complexity of $\mathcal{O}(D^3)$. This makes the metric impractical and inefficient for high-dimensional problems.

\begin{figure}[t]
    \centering
\includegraphics[width=0.49\linewidth]{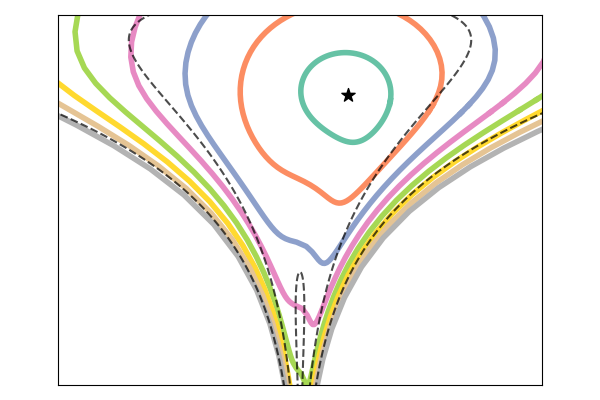}
\includegraphics[width=0.49\linewidth]{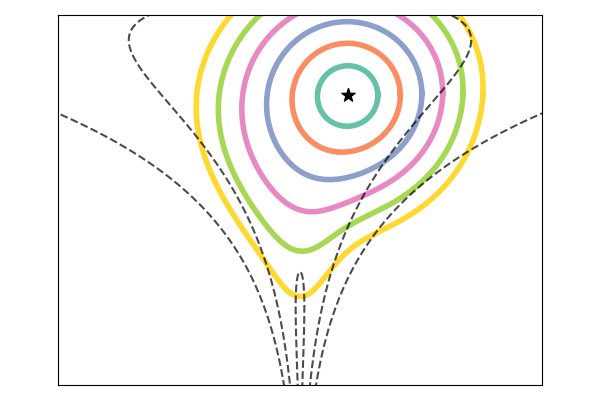}
\caption{    
    Exponential map for Riemannian balls of increasing radius on the Funnel distribution for the Monge metric with $\alpha = 1$ on the left panel. On the right, the plot is analogous but considering the Generative metric with $\lambda=0.1$ and $p_0=0.1$. Each color represents a bigger radius from the base point ($\star$). Both metrics achieve the desired goal, shortening the distances to the points along the narrow funnel that would be difficult to reach in a Euclidean geometry.
    }
    \label{fig:geodesics}
\end{figure}

\paragraph{The Monge Metric}
The computational cost of solving the geodesic equations (Eq.~\ref{eq:geoeqs}) is primarily determined by the inversion of the metric tensor, and consequently metrics with closed-form inverse offer significant savings. The Monge metric by \citet{Hartmann2022} naturally arises from the geometry of the graph of log-density function when viewed as a submanifold embedded in $\mathbb{R}^{D+1}$. Let $\alpha^2 \geq 0$ and $\lambda \geq 0$. The Monge metric and its inverse are given by
\begin{align}
    \label{eq:monge}
    \bG_M(\bx) &= 
    \bI_D
    + \alpha^2 \nabla \ell \nabla \ell^\top,  
    \nonumber \\
    \bG_M^{-1}(\bx) &= 
    \bI_D 
    - \frac{\alpha^2}{1+\alpha^2 \|\nabla \ell\|^2} \nabla \ell \nabla \ell^\top,
\end{align}
where $\ell(x)=\ln p(x)$. As $\alpha^2 \to 0$, the metric reduces to the Euclidean metric $\bI_D$. The determinant required for computing the Hausdorff density (Eq.~\eqref{eq:hausdensity}) is $\det \bG_M(\bx) = 1 + \alpha^2 \|\nabla \ell\|^2$.
Figure~\ref{fig:geodesics} illustrates the exponential map of geodesic balls with increasing radius under the Monge metric. This metric adapts to the geometry of the target distribution, expanding regions based on the local structure of the density.

\paragraph{The Generative Metric}
Another efficient metric, the Generative metric that is proportional to the target density function, was recently proposed by \citet{Kim2024}. One of its advantages is that computing the Christoffel symbols $\Gamma^k_{ij}$ only requires first-order derivatives of the density, whereas the Monge metric (Equation~\ref{eq:monge}) introduces second-order terms. For scalars $p_0 > 0$ and $\lambda \geq 0$, the Generative metric and its inverse are:
\begin{align} \label{eq:gen}
    \bG_g(\bx) &= 
    \left(\frac{p_0 + \lambda}{p(\bx) + \lambda} \right)^2 
    \boldsymbol{I}_D, \\
    \bG_g^{-1}(\bx) &= 
    \left(\frac{p(\bx) + \lambda}{p_0 + \lambda}\right)^2 
    \boldsymbol{I}_D. 
\end{align}
As $\lambda \to \infty$, the metric reduces to the Euclidean metric. The determinant is  
$\det \bG_g(\bx) = \left(\tfrac{p_0 + \lambda}{p(\bx) + \lambda}\right)^{2D}$.
Figure~\ref{fig:fig1} illustrates the effect of $\lambda$ on the Hausdorff density along geodesics $t \mapsto p_{\mathcal{H}}\left(\hat{\gamma}_{(\bx, \bbv)}(t)\right)$, and Figure~\ref{fig:geodesics} again shows how the Generative metric transforms the space.

\subsection{For Bridging the Modes}
\label{sec:metrics_global}

The above metrics adapt for the local curvature and have been designed to improve sampling of, for instance, narrow funnels by re-defining the proximity (see Figure~\ref{fig:geodesics}). For assisting exploration of multimodal targets we need different kinds of metrics: Now we would want a metric that makes modes that are far away in the original Euclidean sense appear closer. 
% (see Figure~\ref{fig:geodesics_multimodal}). 
%
With the exception of the construction of \citet{Lan2014}, which we will discuss in Section~\ref{sec:related_work}, we are not aware of any previous metrics designed for this. Next, we introduce two computationally efficient metrics, with fast inverses and determinants, for assisting multimodal sampling. 

%We start by reminding that any matrix $\bG(\bx)$ defines a valid Riemannian metric as long as it is positive definite for every $\bx \in \man$ and varies continuously on $\man$. Since the inverse of a positive definite matrix is also positive definite, we observe that it is possible to use any of the previously formulated $\bG^{-1}(\bx)$ for defining a metric. 

Geodesic curves maintain a constant velocity norm in the Riemannian sense by construction.  Let $\bx_t=\gamma_{(\bx_0, \bbv_0)}(t)$ be a geodesic curve with velocity $\bbv_t=\dot\gamma_{(\bx_0, \bbv_0)}(t) $, starting from $\bx_0$ with initial velocity $\bbv_0$. If the geodesic moves toward a low-probability region where $p(\bx_t) \to 0$, then the ``mode bridging'' behavior occurs if the Euclidean velocity norms satisfy $\norm{\bbv_0}_2 \ll \norm{\bbv_t}_2$.
This means that as $t$ increases in low-density regions, the geodesics curves accelerate and locally pull the distant modes closer.  

\begin{figure}[t]
    \centering
\includegraphics[width=0.32\linewidth]{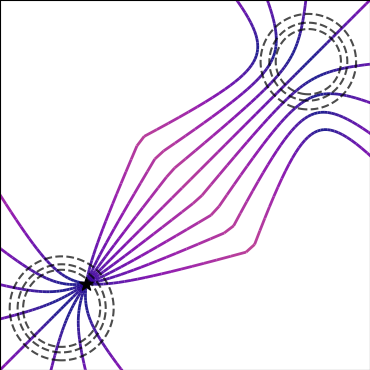}
\hspace{0.8cm}
\includegraphics[width=0.32\linewidth]{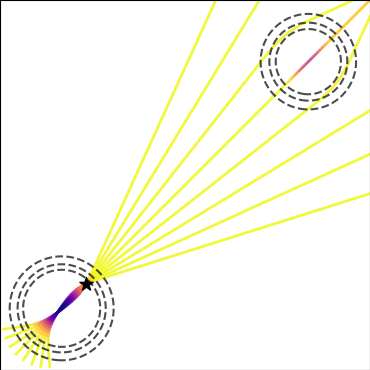}
\caption{
    Effect of the metric for multimodal targets, showing the geodesics (lines) and the relative compression of the distance (color; yellow means faster travel in that area, darker colors mean slower travel). %{\bf Left:} Euclidean metric. 
     {\bf Left:} Inverse Monge metric ($\alpha = 0.001$) helps more the geodesics to reach the other mode, and also slightly compresses the distances in the low-probability region.  
    {\bf Right:} Inverse Generative metric ($\lambda = 1$) compresses the distances in the low-probability region, but twists the paths only slightly.  
    }
    \label{fig:geodesics_multimodal}
\end{figure}

We propose two metrics with the desired behavior, by leveraging the metrics described in Section~\ref{sec:metrics_local} in a novel way. Any matrix $\bG(\bx)$ defines a valid Riemannian metric as long as it is positive definite for every $\bx \in \man$ and varies continuously on $\man$. Since the inverse of a positive definite matrix is also positive definite, we observe that it is possible to use any of the previously formulated $\bG^{-1}(\bx)$ for defining a metric. This gives two new metrics that both help exploring multiple modes in different ways:
%Next, we explain the concrete metrics and their characteristics.
%concretely, we propose:

\paragraph{The Inverse Monge Metric} We use
\begin{align*}
\bG_{IM}(\bx) &= 
    \bI_D 
    - \frac{\alpha^2}{1+\alpha^2 \|\nabla \ell\|^2} \nabla \ell \nabla \ell^\top,
\end{align*}
as the metric tensor, with the inverse $\bG_{IM}^{-1}(\bx) = \bG_M(\bx)$ given by the previously introduced metric tensor of the standard Monge metric
(Eq.~\eqref{eq:monge}). The determinant of this metric is $1/\det(\bG_M)$, and hence it retains the computational efficiency of the original Monge metric.  Figure~\ref{fig:geodesics_multimodal} illustrates the geodesics emanating from one mode of a bimodal distribution under the Inverse Monge metric. The metric twists the curves towards the second mode and slightly increases acceleration (seen by the change of color). Observation~\ref{obs:invmonge} mathematically states the conditions for the change of acceleration caused by the metric.

\begin{observation}
\emph{
    Let $p(\bx)$ be a smooth density function. Let $(\bx_t, \bbv_t)$ be the geodesic flow with initial conditions $(\bx_0,\bbv_0)$  with respect to the Inverse Monge metric, such that $\bx_0$ is a local maximum.
    Then $\norm{\bbv_t}_{2}\geq \norm{\bbv_0}_{2}$ 
    for all $t\neq 0$.
    } \label{obs:invmonge}
\end{observation}

\paragraph{The Inverse Generative Metric} We use
\begin{align*}
\bG_{Ig}(\bx) &= 
    \left(\frac{p(\bx) + \lambda} {p_0 + \lambda}\right)^2 
    \boldsymbol{I}_D,
\end{align*}
and obtain the inverse $\bG_{Ig}^{-1}(\bx) = \bG_g(\bx)$ as the metric tensor of the standard Generative metric (Eq.~\eqref{eq:gen}) and the determinant as $1/\det(\bG_g)$. Again, the computational efficiency of the original Generative metric is retained. Figure~\ref{fig:geodesics_multimodal}  illustrates the main effect of the metric, that is to accelerate on low density regions (indicated by the light color); it also twists the trajectories slightly (best seen within the initial mode and beyond the second mode in the top right corner). 
Additionally Figure~\ref{fig:fig1} illustrates the behavior in a univariate distribution. 
The acceleration behavior is mathematically stated in Observation~\ref{obs:invgen}.

% Next, we explain how the Inverse Monge and Inverse Generative metrics achieve the desired mode-bridging ability. 

% Figure~\ref{fig:fig1} illustrates this effect for the Inverse Generative metric. The following observations collect the previous idea which we discuss further in Appendix~\ref{app:proof_props}:

\begin{observation}
\emph{
    Let $p(\bx)$ be a smooth density function. Let $(\bx_t, \bbv_t)$ be the geodesic flow with initial conditions $(\bx_0,\bbv_0)$ such that $p(x_0)>0$ with respect to the Inverse Generative metric. Then, for $t$ such that $p(\bx_t) \to 0$ we have $\norm{\bbv_t}_{2}>\norm{\bbv_t}_{0}$.
    } \label{obs:invgen}
\end{observation}

The mathematical details for Observations~\ref{obs:invmonge} and \ref{obs:invgen} are given in Appendix~\ref{app:proof_props}.

\section{Experiments}\label{sec:exp}

We evaluate \ourmethod\ for targets with sharp curvature (Section~\ref{sec:exp1}), multiple modes (Section~\ref{sec:exp2}), or both (Section~\ref{sec:exp3}), always considering different choices of the metric. We also empirically quantify the effect of the numerical integrator. A code reproducing the experiments is available at
%\footnote{ Code available at:
\href{https://github.com/williwilliams3/magss}{github.com/williwilliams3/magss}.  
%}.

\paragraph{Evaluation}
We use primarily targets with known reference samples, which allows measuring the accuracy using the 1-Wasserstein (earth mover's) distance with the samples provided by the algorithm \citep{Flamary2021}. 
Besides accuracy, we quantify the samplers with the probability of jumping between the different modes, as the raw ratio of consecutive samples that are within separate modes (defined manually for each problem).

\paragraph{Comparison methods}
To showcase the effect of the metric we will be running \ourmethod\ also with in Euclidean metric, with $\bG(\bx) = \boldsymbol{I}_D$, and we additionally compared against the No-U-Turn Sampler, parallel tempering and diffusive Gibbs sampling.

The No-U-Turn Sampler (NUTS) is an auxiliary-variable sampler that augments the position $\bx_t$ with a velocity $\bbv_t$ which jointly follow the Hamiltonian dynamics \citep{Neal2011}. It adaptively determines the integration time by stopping at the first U-turn, i.e., the first time $t>0$ such that $\langle \bx_t - \bx_0, \bbv_t \rangle < 0$ \citep{Hoffman2014}.

Parallel tempering (PT) runs many Markov Chains in parallel, each of which has $p(\bx)^{1/\tau_i}$ as targets for different temperatures $ \tau_i \geq 1$, with $\tau_1 = 1$ recovering the original target. As $\tau \to \infty$, the density flattens, facilitating transitions between regions of higher densities that are far apart from each other. The parallel chains jumps randomly between each other, thus visiting the modes more often according to a Metropolis-Hastings ratio \citep{Swendsen1986, Geyer1991}. 
Our implementation of PT follows \citet{Latuszynski2025}.

Diffusive Gibbs sampling (DiGS) by \citet{Chen2024} is a sampler designed for addressing multimodality.
It approaches the sampling task by using an auxiliary variable $\tilde{\bx}$ with a Gibbs scheme. It  uses the  variance preserving (VP) \citep{Song2021} noise scaling:  $p(\tilde{\bx}|\bx) = \cN(\tilde{\bx}|\alpha_t \bx, \sigma^2_t)$, where  $\sigma_{t}=\sqrt{1-\alpha_{t}^{2}}$, sampled directly and $p(\bx|\tilde{\bx})\propto p(\tilde{\bx}|\bx) p(\bx)$ sampled through a local MCMC sampler. It has an additional Metropolis within Gibbs proposal scheme $q(\bx|\tilde{\bx}) = \cN(\bx| \tilde{\bx}/\alpha_t, (\alpha_t/\sigma_t)^2)$.  VP has the property that at when $\alpha_t\to 0$ then $p(\tilde{\bx}|\bx)= \cN(\tilde{\bx}|0, \bI_D)$ and when $\alpha_t \to 1$ then, informally, $p(\tilde{\bx}|\bx)= \delta_{\bx}$.

\subsection{Complex unimodal targets}
\label{sec:exp1}

We evaluate the methods on three canonical benchmark targets (funnel, hybrid Rosenbrock and squiggle) which exhibit strong curvature. The densities are given in Appendix~\ref{app:toydist}. Since these targets are unimodal, we only consider the metrics presented in Section~\ref{sec:metrics_local} and exclude PT. 

\begin{figure*}[t]
    \centering
    \includegraphics[width = 0.3\textwidth]{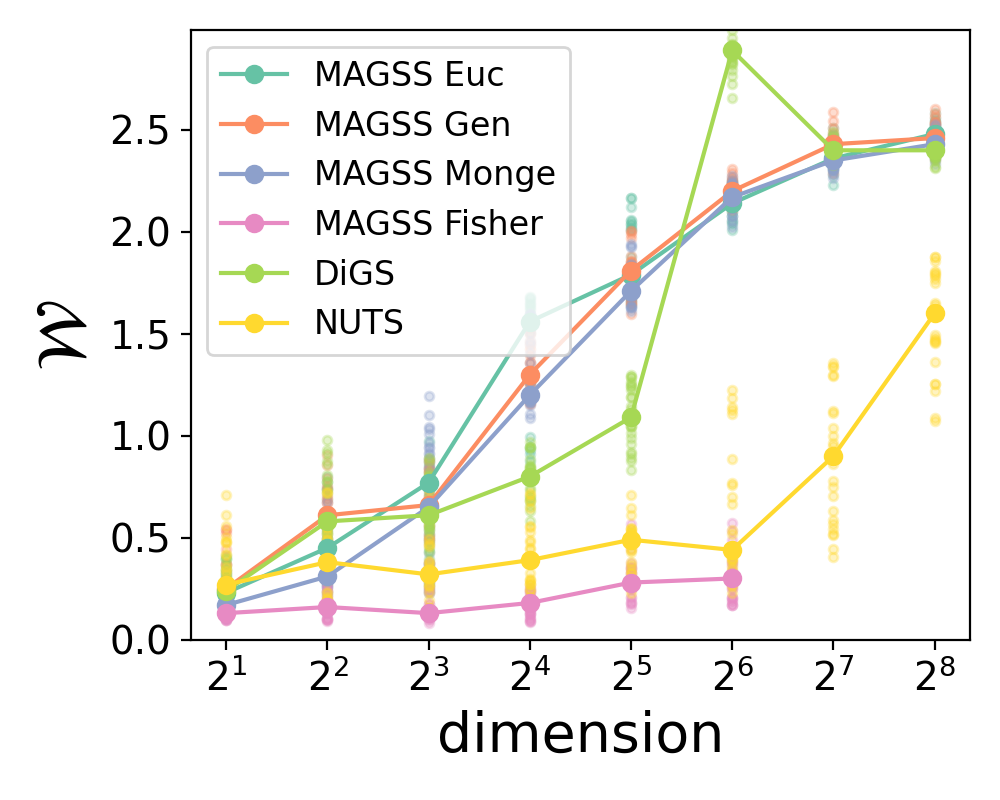}    
    \includegraphics[width = 0.3\textwidth]{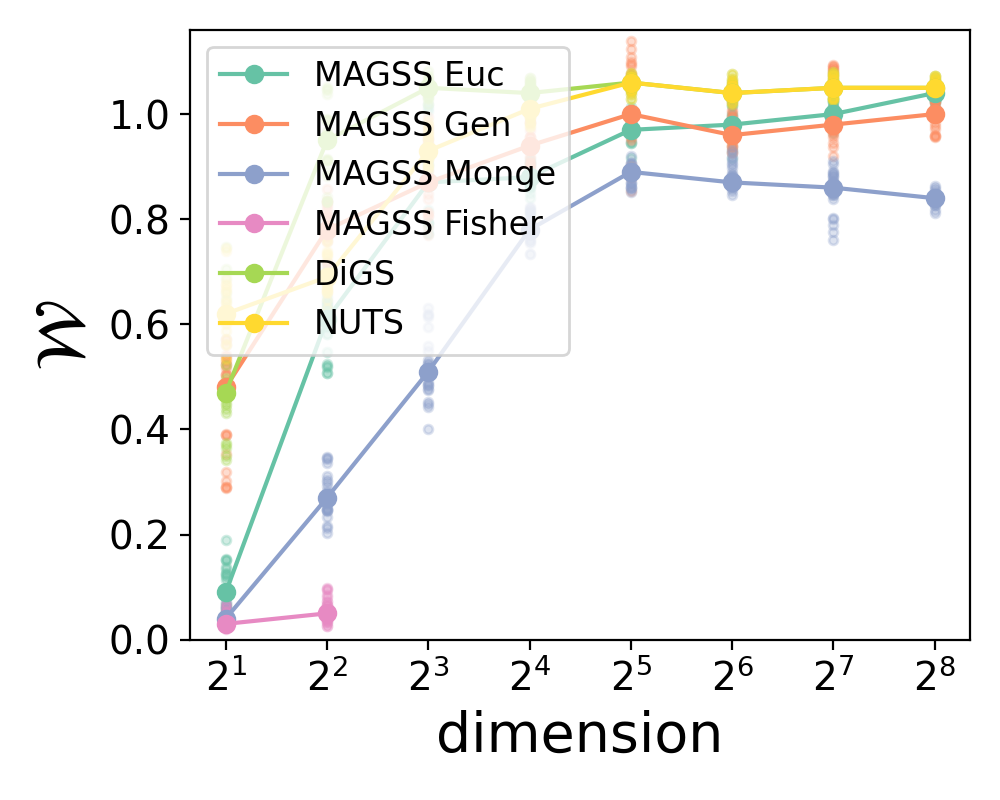}    
    \includegraphics[width = 0.3\textwidth]{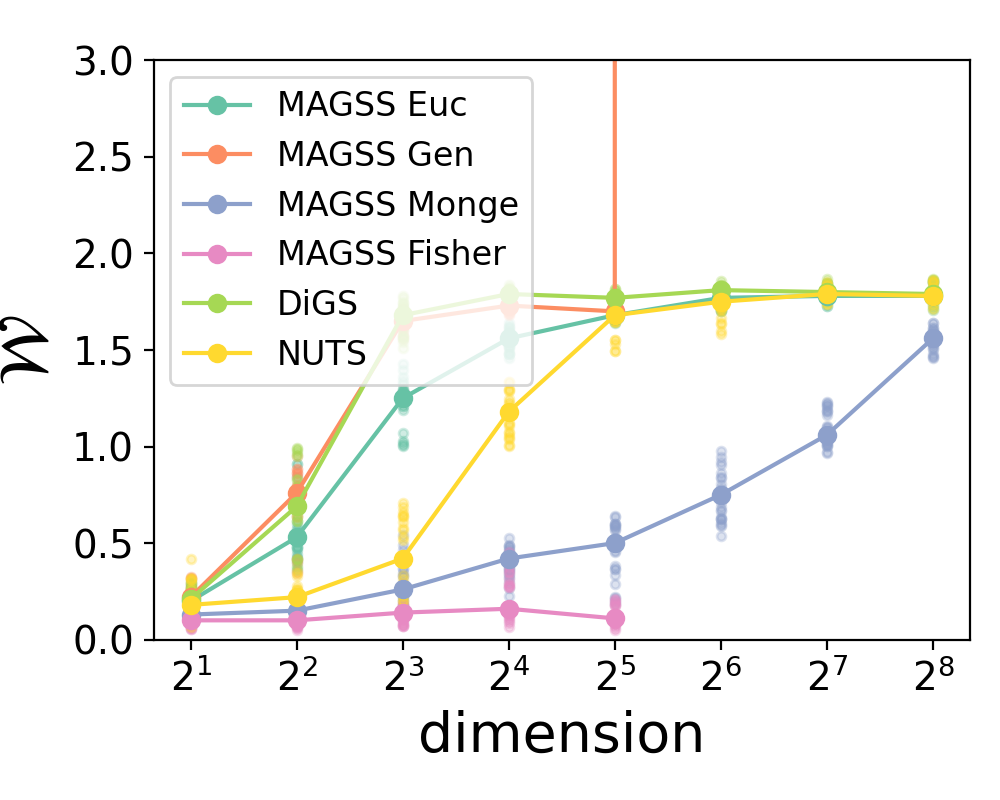}    
    \caption{Univariate sampling accuracy in various metrics (Wasserstein distance, lower is better) for targets of varying dimensionality. The medians over 5 runs are connected with a line.
    Left: Funnel. Middle: Rosenbrock. Right: Squiggle.    
    }
    \label{fig:toydistdims}
\end{figure*}

Figure~\ref{fig:toydistdims} shows that \ourmethod\ with Fisher metric $\bG_F(\bx)$ is clearly superior, but runs out of the limited computational budget already at low dimensions, and the Monge metric $\bG_M(\bx)$ offers notable improvement for Rosenbrock and squiggle targets. DiGS remains on the level of the Euclidean \ourmethod\ and the Generative metric does not help either.

\paragraph{Experiment specification:}
We obtain $10,000$ samples using $10$ chains and omit results for runs that did not complete in 12 hours.
We set $\alpha^2=1$ for the  Monge metric since this value has been shown to work \citep{Hartmann2022}.
We select $\lambda=1$, $p_0=1$  for the  Generative metric without further tuning.
We use Dopri5 integrator with adaptive step-size. We set $w=3$ and $m=8$.  DiGS and NUTS uses a single noise scale $\alpha = 1$ and step-size $0.1$ for MALA within the algorithm.

\subsection{Multimodal with simple modes}
\label{sec:exp2}

For studying mode exploration, we use a target of two
$D$-dimensional Gaussian distributions centered at $-\boldsymbol{1}_{D}$ and $\boldsymbol{1}_{D}$ with scale $\sigma=0.1$ and weights $\{0.2, 0.8\}$.
The distance between the modes ($\sqrt{D}2$) grows for increasing dimensions, making transition between the modes more difficult. Now we only consider the new metrics for boosting mixing between the modes (Section~\ref{sec:metrics_global}).

\begin{figure}[t]
    \centering    
    \includegraphics[width=0.49\linewidth]{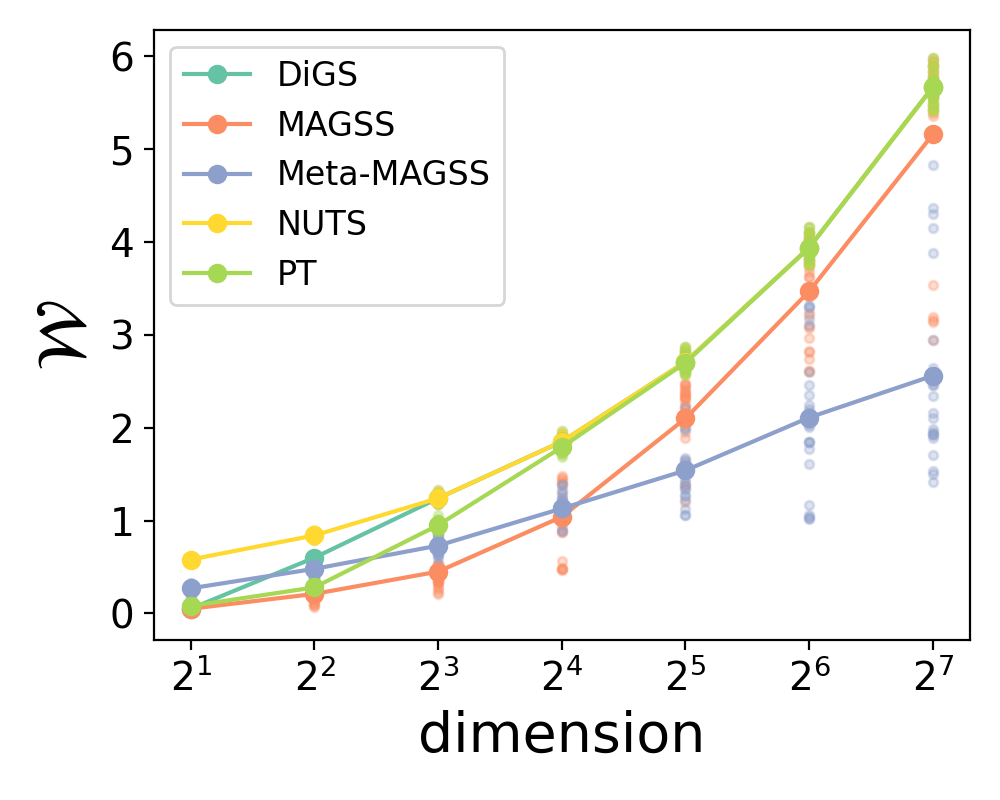}
    \includegraphics[width=0.49\linewidth]{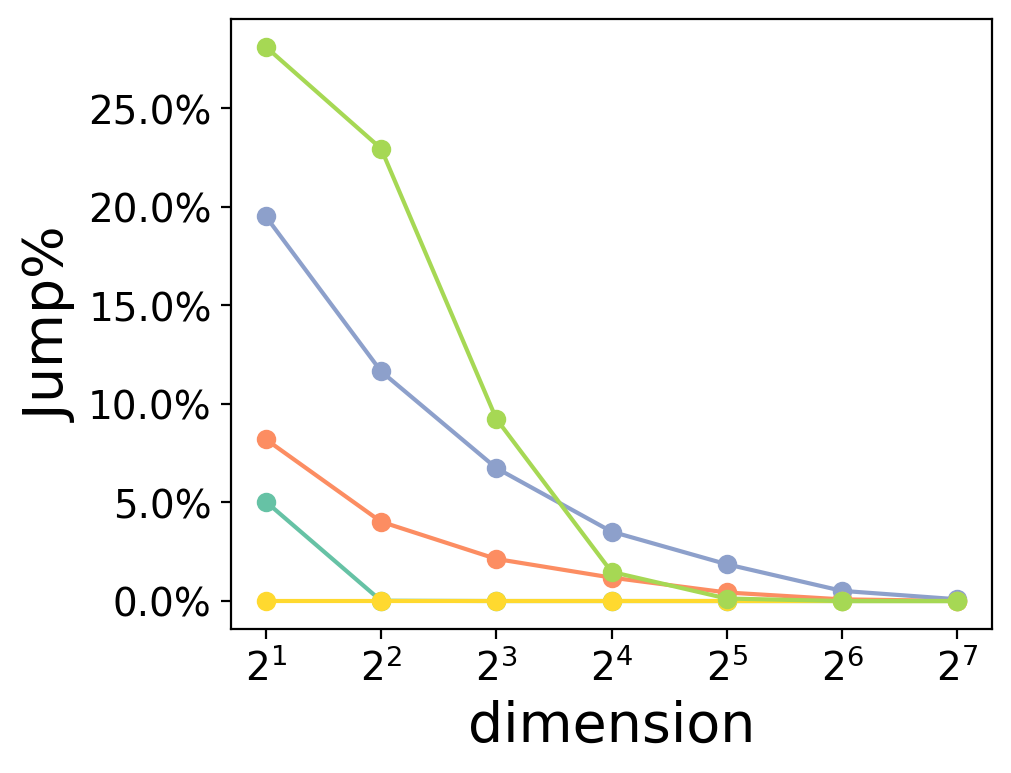}
    \caption{Accuracy (lower is better) for mixture of Gaussians. Both \ourmethod\ variants use $\bG_{IM}$ with $\alpha=0.1$.}
    \label{fig:twogauss}
\end{figure}

Figure~\ref{fig:twogauss} reports the corresponding accuracies and reports the percentage of jumps between the modes. While the comparison methods PT and DiGS explore the modes well in low dimensions, they get completely stuck in one mode for $D \geq 16$. \ourmethod\ and \metaourmethod\ with the Inverse Monge metric ($\alpha=0.1$) are able to jump between the modes even for higher dimensions and the \emph{meta-sampler} is overall the most accurate method.

% {
% \renewcommand{\arraystretch}{0.9} % Reduce row height
% \setlength{\tabcolsep}{3pt} % Reduce column spacing
% \noindent
% \small
% \begin{table}[t]
%     \caption{Mixture of two Gaussians ($\bG_{IM}$, with $\alpha=0.1$).}
%     \centering
%     \begin{tabular}{@{}l *{7}{r} @{}}
%     \toprule
%     jump\% & \multicolumn{6}{c}{dimension} \\
%     \cmidrule(lr){2-7}
%           sampler  & 2  & 4  & 8  & 16  & 32  & 64 \\
%     \midrule
%     \ourmethod    & 8.96  & 5.07  & 2.28  & 0.8   & 0.2   & 0.03  \\
% \metaourmethod & 18.91  & 12.33  & 7.5  & 4.29  & 2.45  & 1.07  \\
%     DiGS    & 5.05  & 0.02  & 0.0   & 0.0   & 0.0   & 0.0   \\
%     PT      & 12.54 & 4.68  & 0.23  & 0.0   & 0.0   & 0.0   \\
%     \bottomrule
%     \end{tabular}
% \label{tab:twogauss}
% \end{table}
% }

\paragraph{Experiment specifications:}For MALA we find a stepsize of $60\%$ acceptance rate for each dimension, since it is close to the optimal for Gaussians \citep{Roberts1998}. For \ourmethod\ we
try the Euclidean metric and the grid: $\alpha^2\in \{10^{-3}, 10^{-2}, 10^{-1}, 1, 10\}$ and $\lambda \in \{10^{-4}, 10^{-3}, 10^{-2}, 10^{-1}, 1, 10\}$. We find $\alpha^2=0.1$ is always optimal. For \metaourmethod\ we fix $\alpha^2=0.1$ based on what was observed for \ourmethod.
We use Dopri5 solver with adaptive step-size. DiGS uses $10$ MALA steps per iterations, $T=100$  equally spaced times between $\alpha_1=10^{-4}$, $\alpha_{1000} = 1-10^{-4}$. PT uses $N=100$ temperatures in the scale $\tau_i = b_{\min}^{-{i}/{N}}$ for $i=1,..,N$ where $b_{\mathrm{min}}=10^{-4}$.

\subsection{Multimodal with complex modes}
\label{sec:exp3}

\begin{figure}[t]
    \centering    
    \begin{minipage}{0.5\linewidth}
        \centering
        \includegraphics[width=\linewidth]{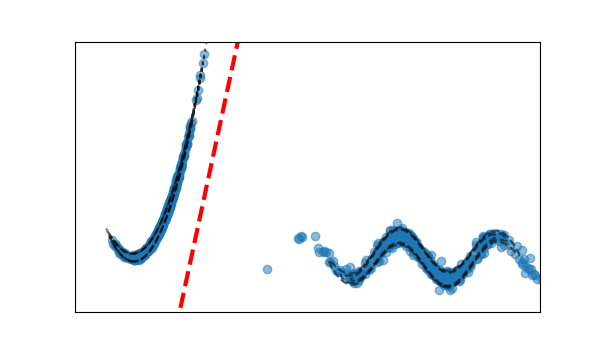}
    \end{minipage}
    \begin{minipage}{0.49\linewidth}
        \centering
        \includegraphics[width=\linewidth]{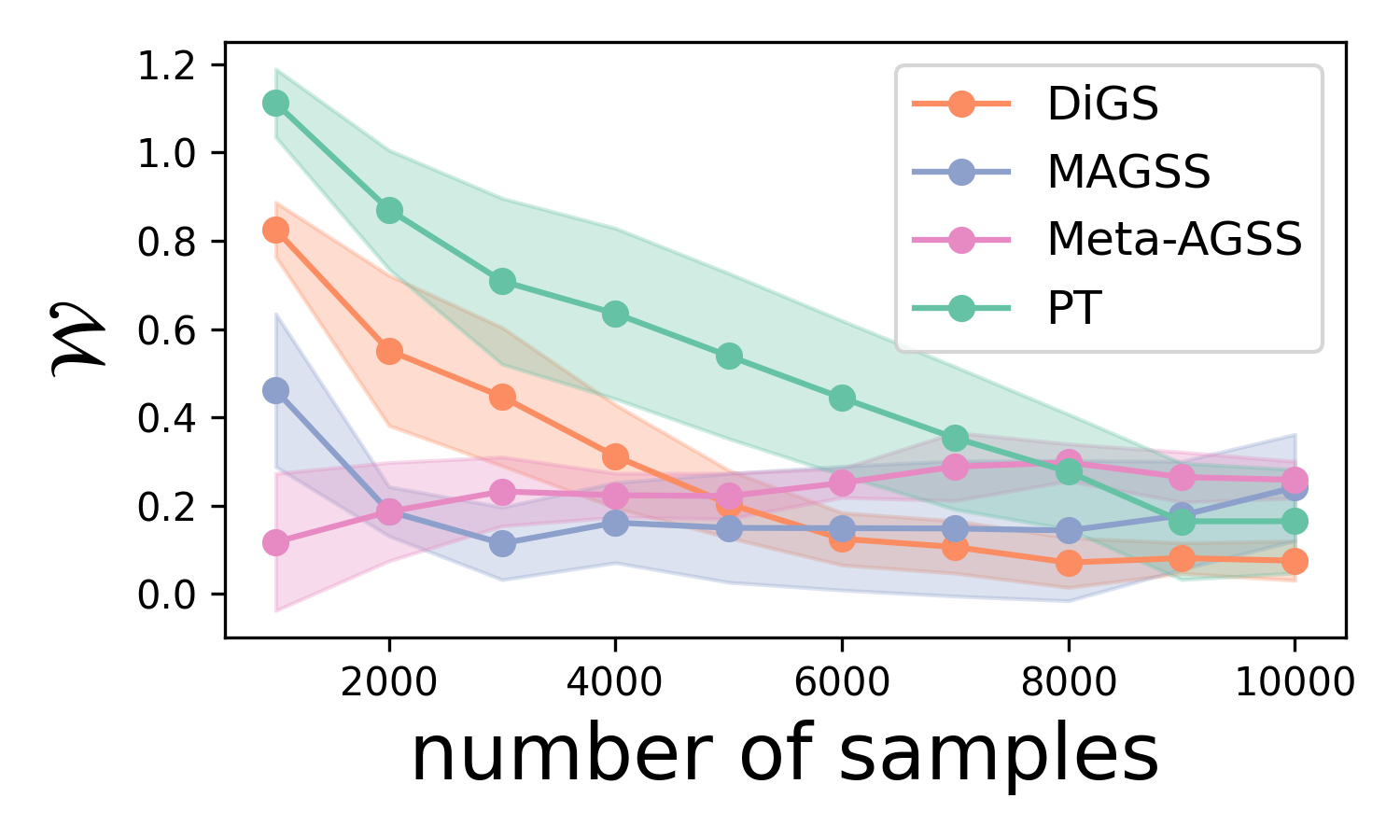}
    \end{minipage}
    \caption{Left: Mixture of narrow distributions, with samples using \metaourmethod.
    Right: Wasserstein distance as a function of iterations (samples).  For \ourmethod\ we use $\bG_{Ig}$ with $\lambda=1$  and  \metaourmethod\ $\bG_{IM}$ with $\alpha^2=10^{-4}$.}
    \label{fig:narrow}
\end{figure}

To demonstrate that we can simultaneously handle multimodality and complex local geometry, we consider a (uniform) mixture of two narrow bivariate distributions, the Rosenbrock and Squiggle distributions (Figure~\ref{fig:narrow} left; the red line is purely for identifying jumps between the modes, Table~\ref{tab:narrow_jumps}).
We use use the Inverse Monge and Inverse Generative metrics.
However, Figure~\ref{fig:narrow} (right) indicates that PT is the least accurate method, requiring substantially more samples for matching the target well. All methods will reach approximately the same Wasserstein distance if ran long enough, but both of our variants achieve it in less samples, confirming more efficient mixing.

\begin{table}[t]
    \centering
    \caption{Mixture of narrow distributions.}
    \begin{tabular}{llll}
\toprule
sampler & metric & jump\% & t(s) \\
\midrule
PT & NA & $6.18$ & 2 \\
DiGS & NA & $0.27$ & 2 \\
\ourmethod & $\bG_{Ig}$, $\lambda=1.0$ & $0.81$ & 178 \\
\metaourmethod & $\bG_{IM}$, $\alpha^2=10^{-4}$ & $2.62$ & 1327 \\
\bottomrule
\end{tabular}
    
    \label{tab:narrow_jumps}
\end{table}

\paragraph{Experiment specifications:}
Obtain $10,000$ samples using $10$ chains.
We run DiGS with 5 noise steps between $0.1$ and $0.9$, and $10$ MALA iterations per sample. PT uses $\tau \in \{1.0, 5.62, 31.62, 177.83, 1000\}$ and thinning of $10$. \ourmethod\ and \metaourmethod\  are tuned using the same grid of values as Experiment~\ref{sec:exp2}, reporting the best based on distances. \metaourmethod\ uses $5$ sweeps and $10$ MALA iterations per sample. PT, DiGS and \metaourmethod\ rely on MALA with stepsize $0.001$ ($\approx60\%$ acceptance rate). We use $w = 3$ and $m=8$ and  the adaptive Dopri8 integrator. %with adaptive step-size. 

\subsection{Field System}

\begin{figure}[t]
    \centering
    \includegraphics[width=0.32\linewidth]{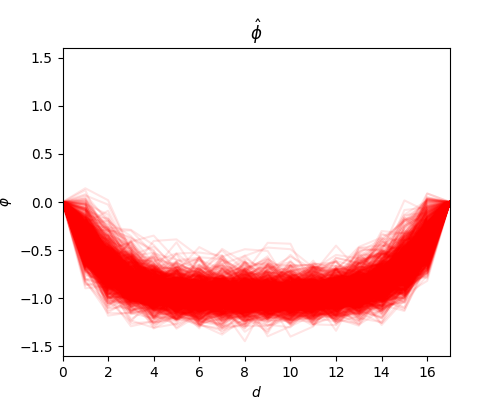}
    \includegraphics[width=0.32\linewidth]{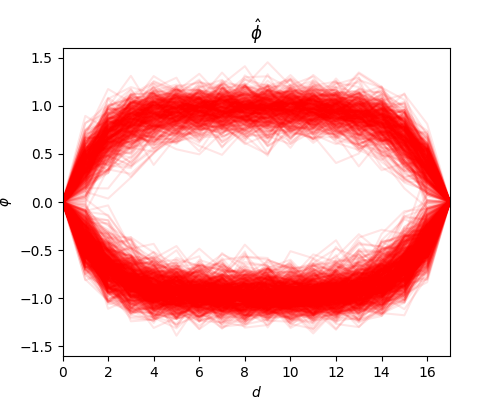}
    \includegraphics[width=0.32\linewidth]{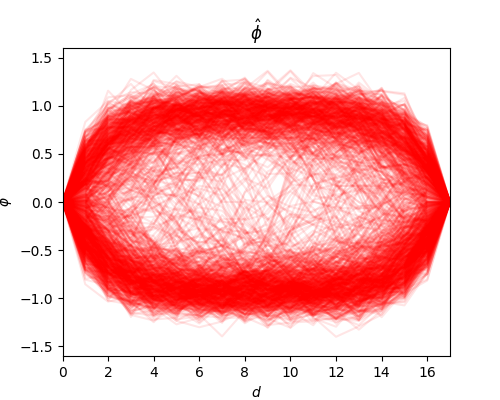}
    \caption{Samples from Allen-Cahn Field System model \citep{Cabezas2024} with $2^{16}$ modes that zig-zag between $-1$ and $1$ on the y-axis over the $D=16$ values at the x-axis. Euclidean DiGS (left) gets stuck in one mode, Parallel Tempering (middle) only explores two dominant modes with constant value over the x-axis, whereas Meta-MAGGS (right; $\boldsymbol{G}_{Ig}$ with $\lambda=10^{-6}$) explores also the modes that switch between the extremes.}
    \label{fig:fieldsystem}
\end{figure}

We include a highly multimodal target distribution of modes ($2^D$ for $D=16$) by replicating the Allen-Cahn Field System model experiment from \citet{Cabezas2024}. The distribution has two global maxima at $(1,..,1)$ and $(-1,..,-1)$, and several lower density modes at points of the form $x_i = \pm 1$ for all $i$. DiGS collapses to a single mode, PT explores only the two most dominant modes, while our \metaourmethod\ explores also the additional  modes (Fig~\ref{fig:fieldsystem}). 

% Since reference samples are not directly obtainable, following \citet{Cabezas2024} we report in Table~\ref{tab:phifour} the  Kernel Stein Discrepancy (KSD V-stat) \citep{Liu2016}. The target density is symmetric along each axis, we initialize the sampling at $(-1,..,-1)$. We choose the dimension $x_8$ as a representative of how well the samplers capture the symmetry. We also report the percentage of samples such that $x_8>0$ the true value is $50\%$. 

The target density is symmetric along each axis. The initial sampling position is $(-1,\ldots,-1)$ and we use the marginal distribution of $x_8$ to evaluate how well each sampler captures the symmetry. In particular, we report the percentage of samples with $x_8 > 0$, which should be $50\%$ under the true distribution. 
Since reference samples are not directly available, we follow \citet{Cabezas2024} and also report the Kernel Stein Discrepancy (KSD V-stat) \citep{Liu2016} in Table~\ref{tab:phifour}.

Our method explores more modes than the competing methods (PT and DiGS),
% This is likely because the sampler's ability to jump between modes results in some samples being located in regions with higher gradient norms, which increases the KSD, since it is depends on gradient of the log target density. It can penalize such transitions even if the overall the mode coverage is better.
although the KSD V-stat is worse. Note, however, that the KSD V-stat does not account for multimodality at all; DiGS has a better value despite covering only a single mode and failing completely in terms of the marginal distribution metric.
% We also note that the KSD V-stat does not account for multimodality: for example, DiGS achieves a low KSD value despite its samples covering only a single mode. 
In contrast, PT and \metaourmethod\ exhibit similar percentages of samples with $x_8 > 0$.
We provide the density of the model, an explanation of the multimodality of the model, and the computation of KSD V-stat in Appendix~\ref{app:toydist}.

\begin{table}[t]
    \centering
    \caption{Field System model}
\begin{tabular}{lllll}
\toprule
sampler &  KSD V-stat & $x_8>0$ & t(s) \\
\midrule
PT &   $0.13\pm0.04$ & $0.35$ & 6 \\
DiGS &   $0.13\pm0.05$ & $0.0$ & 157 \\
META-AGSS  & $2.98\pm0.7$ & $0.33$ & 32 \\
\bottomrule
\end{tabular}    
    \label{tab:phifour}
\end{table}

\paragraph{Experiment specifications:}
We obtain $10,000$ samples using $10$ chains initialized at $(-1,..,-1)$.
DiGS uses $T=1000$  equally spaced times between $\alpha_1=10^{-5}$, $\alpha_{1000} = 1-10^{-5}$. PT uses $N=200$ temperatures in the scale $\tau_i = b_{\min}^{-{i}/{N}}$ for $i=1,..,N$ where $b_{\mathrm{min}}=10^{-5}$. For \metaourmethod\ we try values of $\alpha$ and $\lambda$ in powers of ten, finding $\lambda=10^{-6}$ maximizes the number of jumps between modes and a single sweep. We use $w = 3$ and $m=8$ and  the Dopri5 integrator with adaptive step-size. For all methods  MALA uses $10$ iterations per sample and stepsize $0.005$ (roughly 60\% acceptance rate). 

\subsection{Effect of numerical integrator} \label{sec:gmm}

We use numerical integrators for computing the geodesics in Eq.~\eqref{eq:geoeqs}. To explore the effect of the integrator, we present results for broad range of integrators for a multimodal benchmark task  considered previously by \citet{Chen2024}. The target is a 40-mode Gaussian mixture model with equal weights and each component of variance $\sigma=0.1$ where the means are distributed uniformly on the square $(-40, 40)^2$.

\begin{figure}[t]
    \centering
    \includegraphics[width=0.49\linewidth]{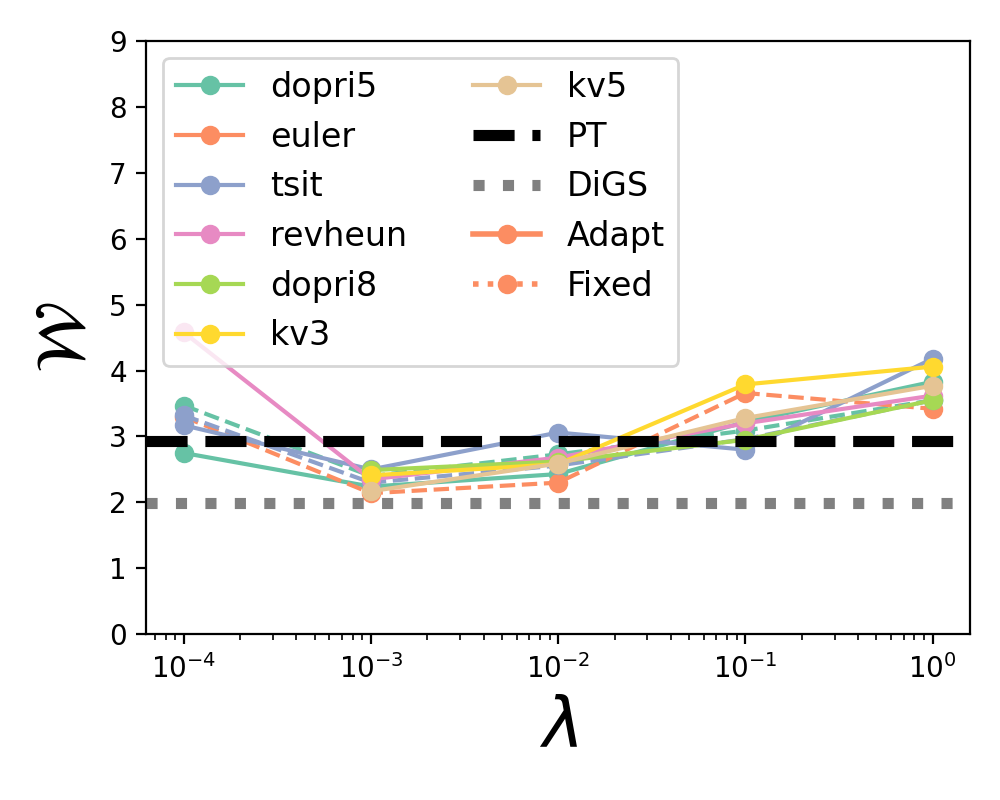}
    \includegraphics[width=0.49\linewidth]{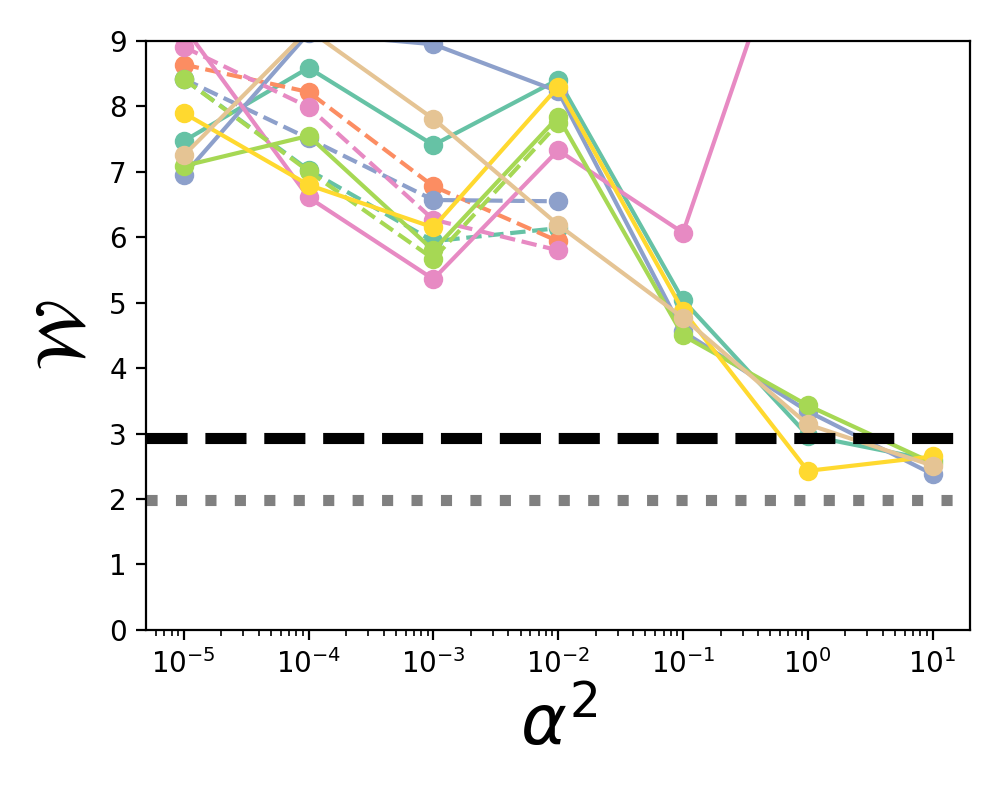}
    \includegraphics[width=0.49\linewidth]{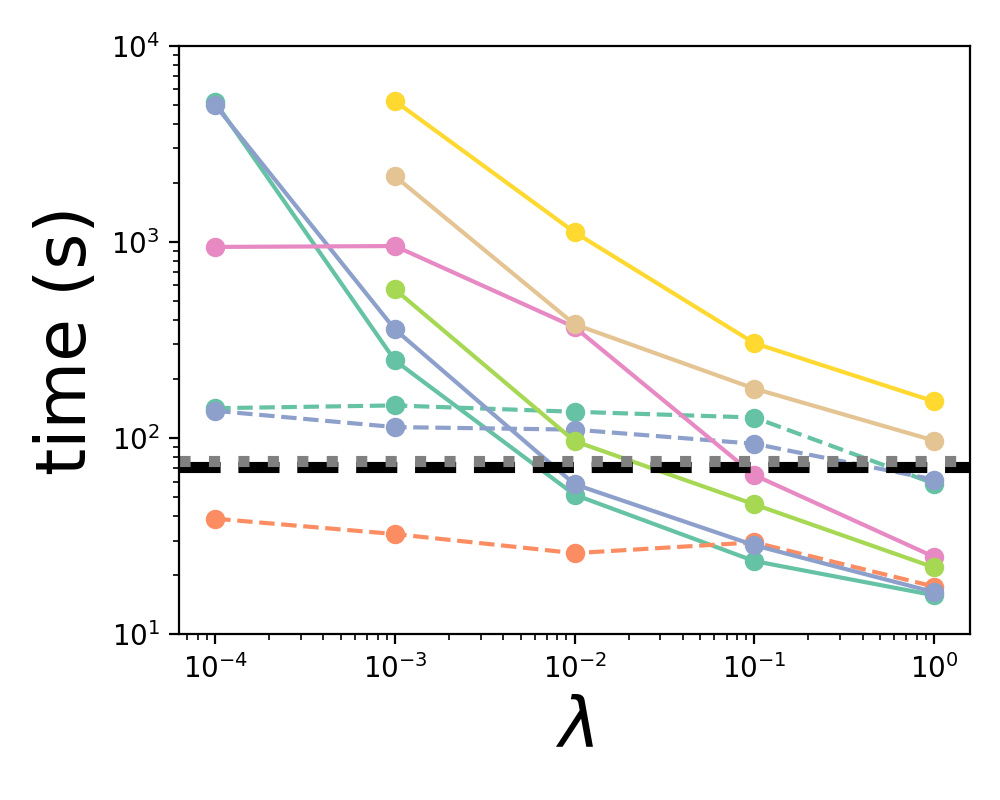}
    \includegraphics[width=0.49\linewidth]{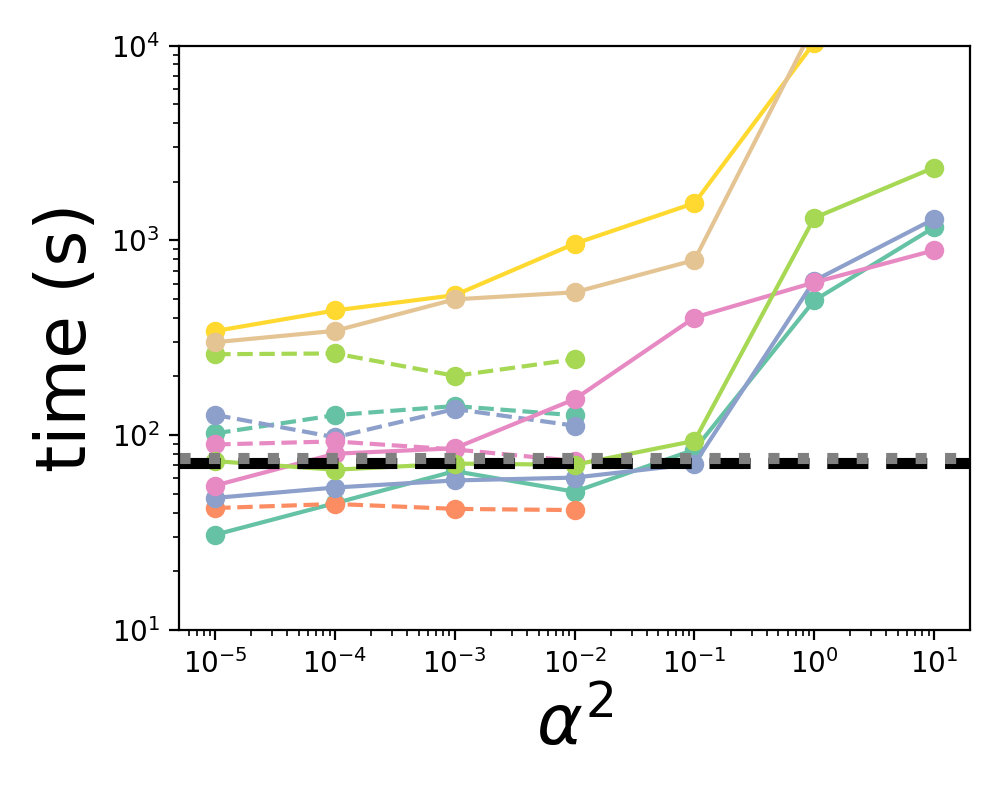}
    \caption{Different numerical integrators for the 40 mode Gaussian mixture model. 
    The black dotted lines are PT and DiGS.
    Left: Inverse Generative metric with parameter $\lambda$. Right: Inverse Monge metric with parameter $\alpha^2$.     
    }
    \label{fig:gmm40}
\end{figure}

We use seven different integrators for $\bG_{IM}$ and $\bG_{Ig}$ metrics, including both adaptive and fixed step-sizes as implemented by \citet{Kidger2021}, and report the results in Figure~\ref{fig:gmm40}. The three main conclusion are: (a) For metrics that are further away from Euclidean (large $\alpha$ or small $\lambda$) the integration time for adaptive methods grows dramatically. This is a natural consequence of operating in a less flat geometry.
(b) For good accuracy we typically need to use such a geometry, which means there is inherent compromise between accuracy an computation.
(c) Simple fixed-step integrators, even the Euler method, are efficient when they work, but for robustness we recommended adaptive methods. We recognize dopri5 as a good practical recommendation, but Euler is worth trying for the Inverse Generative metric.

\paragraph{Experiment specifications:}
Obtain $10,000$ samples using $10$ chains. 
PT has $\tau \in \{1, 5.62, 31.62, 177.83, 1000\}$ and thinning of $200$. DiGS uses $\alpha=0.1$, thinning of $200$ and $5$ MALA steps per step. MALA has step size of $0.1$. 
 \ourmethod\ is run with $w=3$, $m=8$ for the $\bG_{IM}$ and $\bG_{Ig}$ and  metrics with the parameter grid of Experiment~\ref{sec:exp2}. We test seven different numerical integrators of Equation~\eqref{eq:geoeqs}. The fixed integration size is $0.01$. Details in Appendix~\ref{app:exp}.

\section{Related work}
\label{sec:related_work}

\citet{Lan2014} constructed the Wormhole Hamiltonian Monte Carlo where a specific geometry is built to (only) connect the modes of multimodal distributions. The modes are first identified along the Markov Chain evolution. After a new mode identification, it "stores" the mode's location for later use. A jump using the updated mode candidates guarantees correct detailed-balance equations. While this work served as an inspirational motivation for us, it requires notable additional components. However, \ourmethod \ does not require separate identification or storage of the modes, but instead shrinks the distances by naturally warping the space.

\section{Conclusions}

Our aim was to show that local curvature and multimodality can be addressed by the same set of tools, namely Riemannian geometry. We provided a concrete Riemannian slice sampler, introduced two new metrics for improving mixing between modes, and showed that we can achieve accuracy and mixing comparable to recent samplers designed specifically for multimodal targets, by only using  Riemannian metrics for this task.

One obvious limitation is the computational cost, caused by numeric integration of the geodesics. Even when using metrics with fast inverses, the per-iteration cost of \ourmethod\ is larger than of competing methods. However, we note that we used maximally exact solvers rather than seeking for the highest computational efficiency. Now that the principle has been demonstrated, the use of more approximative numerical integrators for speeding up the overall computation could be studied in future work.
% 

% \begin{contributions} % will be removed in pdf for initial submission 
% 					  % (without ‘accepted’ option in \documentclass)
%                       % so you can already fill it to test with the
%                       % ‘accepted’ class option
%     Briefly list author contributions. 
%     This is a nice way of making clear who did what and to give proper credit.
%     This section is optional.

%     BW conceived the idea, developed the method, wrote the code and wrote the paper.
%     HY helped the discussion, wrote PT code and collaborated in the DiGS code.
%     MH helped with the writing, improves proposition 1.
%     HL helped with the theoretical understanding and writing. 
%     GA participated in the discussion. 
%     AK helped with the writing and discussions. 
% \end{contributions}

\begin{acknowledgements} % will be removed in pdf for initial submission,
						 % (without ‘accepted’ option in \documentclass)
                         % so you can already fill it to test with the
                         % ‘accepted’ class option
BW, HY, HPHL and AK were supported by the Research Council of Finland Flagship programme: Finnish Center for Artificial Intelligence (FCAI), and additionally by grants: 363317, 345811 and 348952. GA was supported by the DFF Sapere Aude Starting Grant ``GADL''. The authors wish to acknowledge CSC – IT Center for Science, Finland, for computational resources.
\end{acknowledgements}

% References
\bibliography{bibtex.bib}

\appendix

\newpage
% Add contents to table of contents
\addtocontents{toc}{\protect\setcounter{tocdepth}{2}}

\onecolumn 

\setcounter{figure}{0}
\setcounter{table}{0}
\setcounter{equation}{0}
\setcounter{thm}{0}
\setcounter{proposition}{0}
\setcounter{observation}{0}
\setcounter{assumption}{0}

\title{
Geodesic Slice Sampler for Multimodal Distributions with Strong Curvature
\\(Supplementary Material)}
% \maketitle
\begin{center}
    {\huge \bf Appendix}
\end{center}

\tableofcontents 
% \newpage

\section{\ournamefull } \label{app:agss}

\subsection{Meta-\ournamefull } 
The \metaourmethod found in algorithm~\ref{alg:metaagss} is the combination of \ourmethod\ for $K$-steps followed by a local MCMC sampler for $L$-steps. 
\begin{algorithm}[H]
    \caption{\metaourmethod}
    \label{alg:metaagss}
    \textbf{Input:} Initial position $\bx^{[0]}$ and metric components $\bG(\bx)$. 
    Parameters $m\in \mathbb{N}$, $w\geq 0$, $K$ sweeps, $L$ steps of local MCMC sampler. \\
    \textbf{Output:} $N$ samples $\bx^{[n]}$.
    \begin{algorithmic}[1]
        \For{$n \leftarrow 1, \dots, N$}
            \State Let $\bx \gets \bx^{[n-1]}$
            \For{$k \leftarrow 1, \dots, K$}
                \State Update $\bx$ by \ourmethod\ with initial position $\bx$
            \EndFor
            \For{$l \leftarrow 1, \dots, L$}
                \State Update $\bx$ by local MCMC with initial position at $\bx$.
            \EndFor
            \State Set $\bx^{[n]} \gets  \bx$
        \EndFor
    \end{algorithmic}
\end{algorithm}

\subsection{Step-out and Shrinkage procedures}
The stepping-out and shrinkage procedures are Algorithm~\ref{alg:stepout} and Algorithm~\ref{alg:shrink} respectively, these algorithms are taken from \citet{Durmus2023}. Our code implementation of the step-out procedure has vectorized both while loops in Algorithm~\ref{alg:stepout}. This is done by evaluating the log density on all possible step-out points at once (vectorized). The code implementation of the shrinkage procedure (Algorithm~\ref{alg:shrink}) has a max number of iteration set at $100$ for the while loop, which if exceeded defaults back to the previous point of the chain. In the algorithm boxes we use the notation for the exponential map $ \gamma_{(x,v)}(t)$. JAX is used to handle automatic differentiation and the samplers are coded in the style of Blackjax \citep{Jax2018, Cabezas2024b}.

\begin{algorithm}[H]
\caption{Stepping-out procedure. Call it $\text{Step-out}_{w,m}(s, \gamma_{(\bx, \bbv)})$}\label{alg:stepout}
\textbf{Input:} point $x \in \mathcal{M}$, direction $v \in \mathbb{S}^{d-1}_x$, level $s \in (0, p(x))$, hyperparameters $w \in (0, \infty)$ and $m \in \mathbb{N}$\\
\textbf{Output:} two points $\ell, r \in \mathbb{R}$ such that $\ell < 0 < r$
\begin{algorithmic}[1]
\State Draw $u \sim \text{Unif}([0, w])$.
\State Set $\ell := -u$ and $r := \ell + w$.
\State Draw $\iota \sim \text{Unif}(\{1, \dots, m\})$.
\State Set $i = 2$ and $j = 2$.
\While{$i \leq \iota$ and $p_{\cH}(\gamma_{(x,v)}(\ell)) > s$}
    \State Set $\ell = \ell - w$.
    \State Update $i = i + 1$.
\EndWhile
\While{$j \leq m + 1 - \iota$ and $p_{\cH}(\gamma_{(x,v)}(r)) > s$}
    \State Set $r = r + w$.
    \State Update $j = j + 1$.
\EndWhile
\State \textbf{return} $(\ell, r)$
\end{algorithmic}
\end{algorithm}

\begin{algorithm}[H]
\caption{Shrinkage procedure. Call as $\text{Shrink}_{\ell,r}(s, \gamma_{(\bx, \bbv)})$}\label{alg:shrink}
\textbf{Input:} point $x \in \mathcal{M}$, direction $v \in \mathbb{S}^{d-1}_x$, level $s \in (0, p(x))$ and parameters $\ell < 0 < r$\\
\textbf{Output:} point $\theta \in L(x,v,s) \cap [\ell, r]$
\begin{algorithmic}[1]
\State Draw $\theta_h \sim \text{Unif}((0, r - l))$.
\State Set $\theta := \theta_h - 1_{\{\theta_h > r\}}(r - l)$.
\State Set $\theta_{\min} := \theta_h$.
\State Set $\theta_{\max} := \theta_h$.
\While{$p_{\cH}(\gamma_{(x,v)}(\theta)) \leq s$}
    \If{$\theta_h \in [\theta_{\min}, r - l]$}
        \State Set $\theta_{\min} = \theta_h$.
    \Else
        \State Set $\theta_{\max} = \theta_h$.
    \EndIf
    \State Draw $\theta_h \sim \text{Unif}((0, \theta_{\max}) \cup [\theta_{\min}, r - l))$.
    \State Set $\theta = \theta_h - 1_{\{\theta_h > r\}}(r - l)$.
\EndWhile
\State \textbf{return} $\theta$.
\end{algorithmic}
\end{algorithm}

% \subsection{The Geodesic Equations} \label{app:geoeq}
% We provide only the necessary background on Riemannian geometry. 

% A Riemannian metric tensor is the map $g(\cdot, \cdot):T_{\bx}\man \cross T_{\bx}\man\to \R$ where $T_{\bx}\man$ is the tangent space at position $\bx$ in $\man$. The map must satisfy  positive definiteness for every $\bx \in \man$. Namely, the components $\bG(\bx)$ are such that for all vectors $\bbv, \boldsymbol{u} \in T_{\bx}\man$, then $g(\bbv, \boldsymbol{u}) = \boldsymbol{v}^\top \bG(\bx) \boldsymbol{u} > 0$ (that is, $\bG(\bx)$ is positive definite). 

% Geodesic equations generalize straight lines to curved manifolds. Let $\man$ be a Riemannian manifold with metric tensor $\bG$. The tangent space at a point $\bx$ is denoted by $T_{\bx}\man$. A geodesic curve is determined by solving a second-order differential equation with initial conditions $\bx_0 \in \man$ and $\bbv_0 \in T_{\bx}\man$. The metric components are given by $g_{ij} = \bG(\bx)_{ij}$, and their inverse by $g^{km} = \bG^{-1}(\bx)_{km}$. The geodesic equations are:
% \begin{align*}
%     \dot \bx_k &= \bbv_k, \nonumber \\
%     \dot \bbv_k &= - \norm{\bbv}^2_{\Gamma^k}, \quad \mathrm{for}\ k = 1, \ldots, D. 
% \end{align*}
% The Christoffel symbols, using Einstein summation convention, are: $\Gamma^k_{ij} = \tfrac{1}{2} g^{km} ( \partial_i g_{m j} +  \partial_j g_{i m} - \partial_m g_{i j})$.

\subsection{The Geodesic Equations} \label{app:geoeq}
A Riemannian metric is a smooth, symmetric, and positive-definite tensor $g: T_{\bx}\man \times T_{\bx}\man \to \mathbb{R}$ for each point $\bx \in \man$. In coordinates, the metric is represented by a positive-definite matrix $\bG(\bx)$ such that for all $\bbv, \boldsymbol{u} \in T_{\bx}\man$, 
\begin{equation*}
g(\bbv, \boldsymbol{u}) = \bbv^\top \bG(\bx) \boldsymbol{u}.    
\end{equation*}
Geodesics are curves $\gamma(t)$ on $\man$ that locally minimize distance and generalize straight lines to curved spaces. They solve the geodesic equation, a second-order ODE determined by the metric. Given initial conditions $\gamma(0) = \bx_0 \in \man$ and $\dot{\gamma}(0) = \bbv_0 \in T_{\bx_0}\man$, the geodesic equation in local coordinates is
\begin{equation*}
\ddot{\gamma}^k(t) + \sum_{i,j=1}^D \Gamma^k_{ij}(\gamma(t)) \dot{\gamma}^i(t) \dot{\gamma}^j(t) = 0, \quad \text{for } k = 1, \ldots, D,
\end{equation*}
where $\Gamma^k_{ij}$ are the Christoffel symbols of the second kind, given by
\begin{equation*}
\Gamma^k_{ij} = \tfrac{1}{2} \sum_{m=1}^D g^{km} \left( \partial_i g_{mj} + \partial_j g_{im} - \partial_m g_{ij} \right),
\end{equation*}
with $g_{ij} = \bG(\bx)_{ij}$ and $g^{km} = (\bG^{-1}(\bx))_{km}$.
Alternatively, defining the position-velocity system with $\bx = \gamma(t)$ and $\bbv = \dot{\gamma}(t)$, the geodesic equations can be expressed as a first-order system (Equation~\ref{eq:geoeqs}):
\begin{align*}
    \dot \bx_k &= \bbv_k, \\
    \dot \bbv_k &= - \norm{\bbv}^2_{\Gamma^k}\quad \mathrm{for}\ k = 1, \ldots, D,
\end{align*}
where $\norm{\bbv}^2_{\Gamma^k}=\sum_{i,j=1}^D \Gamma^k_{ij}(\bx) \bbv_i \bbv_j$.

% The paper introduces the concept of Hausdorff density to account for the change in measure from Euclidean space to the manifold. Can you provide more intuitive explanation on why this adjustment is important for maintaining proper sampling behavior?

\subsection{Sampling Uniformly from the Unit Tangent Sphere}

Recall the unit tangent sphere is defined by
\begin{equation*}
    \dS^{D-1}_g(\bx):=\{\bbv\in \R^D: \ \norm{\bbv}_g^2=1\}.
\end{equation*}
\citet[Appendix C.4]{Durmus2023}  justify the existence of the uniform distribution distribution over $\dS^{D-1}_g(\bx)$. One method for producing samples from the uniform distribution on the unit tangent sphere is:
\begin{enumerate}
    \item Sample $ \boldsymbol{z} \sim N(0, \bI) $.
    \item Transform $ \boldsymbol{v} \gets \boldsymbol{G}^{-\tfrac{1}{2}}(\boldsymbol{x}) \boldsymbol{z}$, then $\bbv$ is distributed according to $\mathcal{N}(0, \boldsymbol{G}^{-1}(\boldsymbol{x}))$.
    \item Compute the Riemannian norm $ \|\boldsymbol{v}\|_g = \sqrt{\boldsymbol{v}^T \bG(\bx) \boldsymbol{v}} $.
    \item Project to the boundary  $\boldsymbol{v} \gets \frac{\boldsymbol{v}}{\|\boldsymbol{v}\|_g}$. 
\end{enumerate}

% \paragraph{Rational behind the transformation}
% The initial vector is $\boldsymbol{z} \sim N(0, \boldsymbol{I})$, then the transformation $ \boldsymbol{v} = \boldsymbol{G}^{-\tfrac{1}{2}}(\boldsymbol{x}) \boldsymbol{z}$ is such that the norm in the tangent space is,
% \begin{align*}
%     \norm{\boldsymbol{v}}_{\boldsymbol{G}(\boldsymbol{x})} &=  \boldsymbol{v}^\top \boldsymbol{G}(\boldsymbol{x}) \boldsymbol{v} \\
%                     &=  (\boldsymbol{G}^{-\tfrac{1}{2}}(\boldsymbol{x}) 
%                      \boldsymbol{z}
%                      )^\top \boldsymbol{G}(\boldsymbol{x}) 
%                      (\boldsymbol{G}^{-\tfrac{1}{2}}(\boldsymbol{x}) \boldsymbol{z}  )\\
%                      &= \boldsymbol{z}^\top \boldsymbol{z}. \\
% \end{align*}
% That is, the metric from the original space is preserved. 

% \subsection{The Hausdorff measure}  \label{app:hausdorff}

% The volume form of a Riemannian Manifold with metric $\bG(\bx)$ is defined as $V(\dbx) := \sqrt{\det\bG(\bx)}\dbx$.
% For technical details about the volume form, interested readers can consult Proposition 2.41 in \citet{Lee2018}.
% The volume element gives the natural measure on the manifold, analogous to the Lebesgue measure in the Euclidean space \citep{Durmus2023}. The Hausdorff density is defined as the density which 
% integrates to one with respect to the volume element
% \begin{equation*}
%     p_{\cH}(\bx) = \frac{ p(\bx)}{\sqrt{ \det \bG(\bx)}}  .
% \end{equation*}

\subsection{The Hausdorff measure}  \label{app:hausdorff}

The volume form of a Riemannian manifold with metric $\bG(\bx)$ is defined as $V(\dbx) := \sqrt{\det\bG(\bx)}\dbx$.
For technical details about the volume form, interested readers can consult Proposition 2.41 in \citet{Lee2018}.
The volume element gives the natural measure on the manifold, analogous to the Lebesgue measure in Euclidean space \citep{Durmus2023}. The Hausdorff density is defined as the density which 
integrates to one with respect to the volume element:
\begin{equation*}
    p_{\cH}(\bx) = \frac{ p(\bx)}{\sqrt{ \det \bG(\bx)}}  .
\end{equation*}

An intuitive explanation for the volume element can be thought in terms of change-of-variables in Euclidean space. When transforming coordinates via a diffeomorphism $\phi: \mathbb{R}^D \to \mathbb{R}^D$, the standard density must be adjusted by the Jacobian determinant to preserve probability mass. That is, $|\det J|$ accounts for local volume distortion.

When $\phi:\mathbb{R}^D \to \mathbb{R}^d$, where $d>D$ maps from a lower-dimensional Euclidean space onto a manifold embedded in higher dimensions, the Jacobian $J$ is generally rectangular. In this case, the induced Riemannian metric (Pullback metric) on the manifold is $\bG(\bx) = J J^\top$, and the volume change is given by $\sqrt{\det \bG(\bx)}$, which generalizes $|\det J|$.

Thus, the Hausdorff density $p_{\cH}$ adjusts the density $p$ to be properly normalized on the manifold with respect to the intrinsic geometry. This adjustment ensures correct sampling and integration as seen in Proposition~\ref{prop:prop_haus}.

\subsection{Observations and Proposition } \label{app:proof_props}

Denote by $\bG_{IM}(\bx)$ the inverse Monge metric and by $\bG_{Ig}(\bx)$ the inverse generative metric.
The metrics are defined as:
\begin{equation*}
    \bG_{Ig}(\bx) = \left(\frac{p(\bx) + \lambda}{p_0 + \lambda}\right)^2 \bI, \quad
    \bG_{IM}(\bx) = \bI - \frac{\alpha^2}{1 + \alpha^2 \norm{\nabla \ell(\bx)}^2} \nabla \ell(\bx) \nabla \ell(\bx)^\top.
\end{equation*}

\begin{observation}
\emph{
    Let $p(\bx)$ be a smooth density function. Let $(\bx_t, \bbv_t)$ be the geodesic flow with initial conditions $(\bx_0,\bbv_0)$ such that $p(x_0)>0$ with respect to the Inverse Generative metric. Then, for $t$ such that $p(\bx_t) \to 0$ we have $\norm{\bbv_t}_{2}>\norm{\bbv_t}_{0}$.
    } 
\end{observation}

\paragraph{Analysis for the Inverse Generative metric}
Assume a geodesic curve starting at $(\bx_0,\bbv_0)$ satisfies $p(\bx_0) \geq p(\bx_t)$ for all $t \geq 0$ and $p(\bx_t) \to 0$. Recall that along a geodesic curve, the magnitude of the velocity with respect to the metric remains constant:
\begin{equation*}
\norm{\bbv_t}^2_{\bG_{Ig}} = \norm{\bbv_0}^2_{\bG_{Ig}} \quad \forall t.    
\end{equation*}
Thus, the equality holds:
\begin{align*}
    \norm{\bbv_t}^2_{\bG_{Ig} } &= \norm{\bbv_0}^2_{ \bG_{Ig} } \\
    \left(\frac{p(\bx_0) + \lambda}{p_0 + \lambda}\right)^2 \norm{\bbv_0}^2 &= \left(\frac{p(\bx_t) + \lambda}{p_0 + \lambda}\right)^2 \norm{\bbv_t}_2^2 \\
    \left(\frac{p(\bx_0) + \lambda}{p(\bx_t) + \lambda}\right)^2 \norm{\bbv_0}^2 &= \norm{\bbv_t}_2^2.
\end{align*}

Since $p(\bx_0) \geq p(\bx_t)$, it follows that $\norm{\bbv_t}_2^2 \geq \norm{\bbv_0}_2^2$, and as $p(\bx_t)\to 0$ the quantity is arbitrary large. 

\begin{observation}
\emph{
    Let $p(\bx)$ be a smooth density function. Let $(\bx_t, \bbv_t)$ be the geodesic flow with initial conditions $(\bx_0,\bbv_0)$  with respect to the Inverse Monge metric, such that $\bx_0$ is a local maximum.
    Then $\norm{\bbv_t}_{2}\geq \norm{\bbv_0}_{2}$ 
    for all $t\neq 0$.
    } 
\end{observation}

% {\color{red} Disclaimer: In progress}

% \paragraph{Analysis for the Inverse Monge Metric}
% Assume $t\in I$ as described before. Denote by $\theta_t = \arccos \tfrac{\inp{\bbv_t}{\nabla\ell(\bx_t)}}{\norm{\bbv_t}_2\norm{\nabla\ell(\bx_t)}_2}$.
% The Riemannian norm over the geodesic trajectory is preserved, thus:
% \begin{align*}
%     \norm{\bbv_0}_{ \bG_{IM} }^2 &= \norm{\bbv_t}^2_{ \bG_{IM} } \\
%     \norm{\bbv_0}_2^2 - \frac{\alpha^2}{1 + \alpha^2 \norm{\nabla \ell(\bx_0)}^2} \inp{\nabla \ell(\bx_0)}{\bbv_0}^2 &= \norm{\bbv_t}_2^2 - \frac{\alpha^2}{1 + \alpha^2 \norm{\nabla \ell(\bx_t)}^2}_2 \inp{\nabla \ell(\bx_t)}{\bbv_t}^2 \\
%     \norm{\bbv_0}_2^2 \left(1-\frac{\alpha^2 \norm{\nabla\ell(\bx_0)}^2 \cos^2 \theta_0}{1 + \alpha^2 \norm{\nabla \ell(\bx_0)}^2}  \right)&= \norm{\bbv_t}_2^2\left(1-\frac{\alpha^2 \norm{\nabla\ell(\bx_t)}^2 \cos^2 \theta_t}{1 + \alpha^2 \norm{\nabla \ell(\bx_t)}^2}  \right).
% \end{align*}

% Using the equality $\cos^2 \theta_t < \cos^2 \theta_0$ then $-\cos^2 \theta_0 < -\cos^2 \theta_t$. And using $\norm{\nabla\ell(\bx_t)}_2 > \norm{\nabla\ell(\bx_0)}_2$ then 
% \begin{equation*}
% \frac{\alpha^2 \norm{\nabla\ell(\bx_t)}^2}{1 + \alpha^2 \norm{\nabla\ell(\bx_t)}^2} <    
% \end{equation*}

% Therefore in regions where $v_t$ aligns with the gradient $\cos\theta_t$ approaches $1$ and for sufficiently large $\norm{\bbv_t}_2$ the right hand term approaches zero, then  $\norm{\bbv_t}_2^2 > \norm{\bbv_0}_2^2$. 

\paragraph{Analysis for the Inverse Monge metric}

We consider geodesics emanating from a mode. Let $\bx_0$ be a mode, meaning that $\nabla \ell(\bx_0) = 0$. Consider a geodesic emanating form $\bx_0$ with velocity $\bbv_0$, it holds
\begin{align*}
\norm{\bbv_0}_2^2 - \frac{\alpha^2}{1 + \alpha^2 \norm{\nabla \ell(\bx_0)}^2} \inp{\nabla \ell(\bx_0)}{\bbv_0}^2 &= \norm{\bbv_t}_2^2 - \frac{\alpha^2}{1 + \alpha^2 \norm{\nabla \ell(\bx_t)}^2}_2 \inp{\nabla \ell(\bx_t)}{\bbv_t}^2,
\end{align*}
or
\begin{align*}
    \norm{\bbv_0}_2^2+\frac{\alpha^2}{1 + \alpha^2 \norm{\nabla \ell(\bx_t)}^2}_2 \inp{\nabla \ell(\bx_t)}{\bbv_t}^2 = \norm{\bbv_t}_2^2.  
\end{align*}
We see that $\Vert \bbv_t \Vert^2 \geq \Vert \bbv_0\Vert^2$, so any geodesic starting from $\bx_0$ always has a "shrinkage" behavior. So this metric helps bring the entire space close to $\bx_0$ along geodesics, as it shrinks the space towards (multiple) modes. Also note that the collapsing towards modes depends on how flat the region is and how well-aligned the velocity and the gradient of $\ell$ is. For complicated distributions, the behavior should not depend monotonically on $\alpha$.

\textbf{Note:} For a multimodal distribution of dim $\geq 2$, Observation~\ref{obs:invmonge} and Observation~\ref{obs:invgen} guarantee that low-density/increasing-gradient regions the speed increases, but we do not have the guarantee that the geodesics given by the inverse metrics will reach the other modes. The geodesic could twist before reaching the other modes, which could negate the ``teleport/move fast" effect. 

% TODO: Give a detailed explanation here
% Note that this analysis also holds for the inverse Monge metric when the geodesic curve approaches a local maximum of the distribution, which explains the behavior observed in Figure~\ref{fig:hauspdf_inv} around the local maximum values when $\alpha=0.1$.

\begin{figure}[t]
    \centering
    \includegraphics[width=0.4\linewidth]{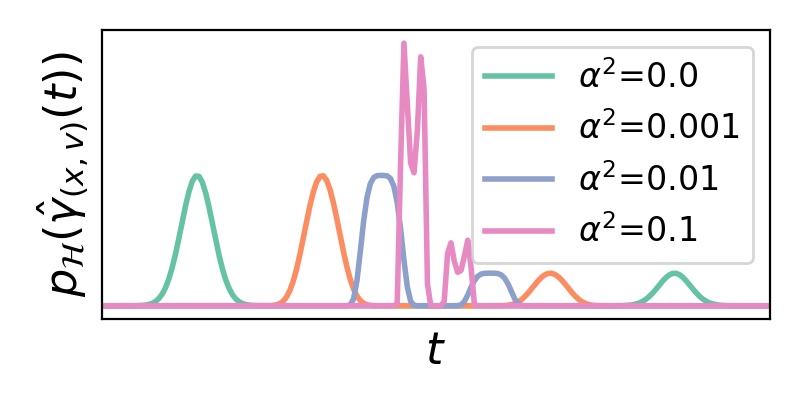}
    \includegraphics[width=0.4\linewidth]{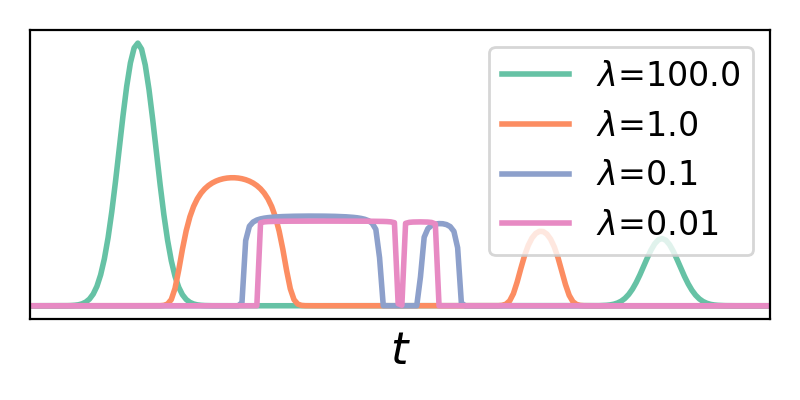}    
    \caption{
    The Hausdorff density of a mixture of two Gaussian distributions evaluated 
    along the geodesic, namely $t\mapsto p_{\cH}(\hat{\gamma}_{(\bx, \bbv)}(t))$ for different values of $\lambda$ and $\alpha^2$. 
    Left: inverse Monge metric. Right: inverse Generative metric}
    \label{fig:hauspdf_inv}
\end{figure}

\begin{proposition} \label{prop:prop_haus}
\emph{
     An MCMC sampler targeting the Hausdorff density on a Riemannian manifold $\mathcal{M}$ with metric tensor $\bG(\bx)$ also targets the correct distribution on the Euclidean space.
     }
\end{proposition}

For a general setting proof of the Proposition  consult Section XII.1, Proposition 1.5 in \citet{Amann2005}.

The volume element on the manifold is defined as $V(\dbx) = \sqrt{\det \bG(\bx)} \dbx$, where $\bG(\bx)$ is the Riemannian metric tensor. Let $\bX$ be a random variable on $\mathcal{M}$ whose law is $p_{\mathcal{H}}$ where $p_{\mathcal{H}}(\bx)$ is the Hausdorff density. Let $B \in \mathcal{B}(\mathcal{M})$ be a Borel set on the manifold $\mathcal{M}$. The probability of $\bX$ being in $B$, under the Hausdorff target density, is given by 
\begin{align} 
    \label{eq:pH}
    \mathbb{P}(\bX \in B) &= \int_{B} p_{\mathcal{H}}(\bx) V(\dbx).
\end{align}
Substituting $p_{\mathcal{H}}(\bx) = \frac{p(\bx)}{\sqrt{\det \bG(\bx)}}$,
\begin{align}
    \label{eq:pE}
    \mathbb{P}(\bX \in B) &= 
    \int_{B} \frac{p(\bx)}{\sqrt{\det G(\bx)}} \sqrt{\det \bG(\bx)} \dbx 
    \nonumber
    \\
    &= \int_{B} p(\bx) \dbx.
\end{align}
Thus, the integral of the Hausdorff density with respect to the volume element on the manifold coincides with the integral of the Euclidean density $p(\bx)$ over the same set $B$.

Since the probabilities computed for any $B \in \mathcal{B}(\mathcal{M})$ are identical whether using \eqref{eq:pH} or \eqref{eq:pE}, the corresponding estimators for the probabilities also coincide. Consequently, an MCMC sampler on the manifold targeting the Hausdorff density $p_{\mathcal{H}}(\bx)$ correctly targets the Euclidean density $p(\bx)$ in $\mathbb{R}^D$.

\section{Additional Experimental Results} \label{app:exp}

\subsection{Logistic regression} \label{app:logreg}
Here we denote by $\theta$ the random variable of interest and by the $x$ input data.
The Logistic regression model \citep{ Girolami2011} is
\begin{equation*}
    p(\by_i | \btheta, \bx_i) = \textrm{Bernoulli}(\by_i|s( \bx_i^\top\btheta )), \quad
    p(\btheta) = \cN(\btheta |0,\alpha \bI_D ), \quad i=1,..,N,
\end{equation*}
where $\alpha=100$ and $s(\cdot)$ is the Sigmoid function. The Fisher Information Metric for this probabilistic model including the addition of the the Hessian of the prior is: $\bG(\bx) = \bX^{\top}\bLambda\bX + \alpha^{-1}\bI $. Where $\bX$ is the covariate matrix and $\bLambda$ is a diagonal matrix with entries $\bLambda_{nn} = s(\bx_i^\top \btheta)\big(1-s(\bx_i^\top \btheta)\big)$.

References samples used for computing the Wasserstein distance are obtained with HMC-NUTS. The samples obtained with the Euclidean and Fisher metrics are just as close to the samples, but Fisher and Monge have higher effective sample size (ESS) and use less shrinkage iterations than the Euclidean metric (See Table~\ref{tab:logreg}). The Monge metric uses the parameter $\alpha=1$. $\mathcal{W}$ is the earth mover's distance. The notation used is $\mathrm{mean} \pm \mathrm{std}$ over 5 runs with different seeds. 

\subsection{Numerical integrators}

The numerical solvers we consider are part of the \texttt{diffrax} package 
\citep{Kidger2021}. We consider three groups of solvers. Simple solvers (euler, tsit, dopri). Implicit solvers (kv). And reversible solvers (revheun). The solvers have the following characteristics:
\begin{itemize}    
    \item euler: The Euler solver can only be used with a fixed step-size.
    \item tsit: Tsitouras' 5/4 method can be used with both fixed and adaptive step-size.
    \item dopri5: Dormand-Prince's 5/4 method can be used with both fixed and adaptive step-size.
    \item dopri8: Dormand-Prince's 8/7 method can be used with both fixed and adaptive step-size. 
    \item kv3: Kvaerno's 3/2 method is an implicit solver can be only used with adaptive step-size.
    \item kv5: Kvaerno's 5/4 method is an implicit solver can only be used  adaptive step-size.
    \item revheun: Reversible Heun method can be used with both fixed and adaptive step-size. 
\end{itemize}

\begin{table}[t]
    \centering
\begin{tabular}{llllllll}
\toprule
model & metr & $\mathcal{W}$ & min ESS & avg ESS & avg step-out & avg shrinkage & t(s) \\
\midrule
aus & euclidean & 0.74 $\pm$ 0.12 & 18 $\pm$ 5 & 228 $\pm$ 12 & 0.14 $\pm$ 0.0 & 3.33 $\pm$ 0.01 & 8.2 \\
 & fisher & 0.58 $\pm$ 0.01 & 177 $\pm$ 9 & 270 $\pm$ 13 & 0.95 $\pm$ 0.0 & 1.0 $\pm$ 0.01 & 3298.6 \\
ger & euclidean & 0.49 $\pm$ 0.01 & 35 $\pm$ 10 & 119 $\pm$ 9 & 0.08 $\pm$ 0.0 & 4.19 $\pm$ 0.02 & 15.8 \\
 & monge & 0.75 $\pm$ 0.02 & 80 $\pm$ 5 & 4673 $\pm$ 113 & 3.97 $\pm$ 0.03 & 0.28 $\pm$ 0.01 & 4665.4 \\
 & fisher & 0.49 $\pm$ 0.0 & 85 $\pm$ 24 & 169 $\pm$ 5 & 0.95 $\pm$ 0.0 & 0.99 $\pm$ 0.01 & 19684.4 \\
hrt & euclidean & 0.63 $\pm$ 0.01 & 137 $\pm$ 26 & 254 $\pm$ 16 & 0.19 $\pm$ 0.01 & 2.72 $\pm$ 0.02 & 7.3 \\
 & monge & 0.74 $\pm$ 0.03 & 394 $\pm$ 55 & 2979 $\pm$ 266 & 2.99 $\pm$ 0.04 & 0.38 $\pm$ 0.02 & 637.4 \\
 & fisher & 0.64 $\pm$ 0.01 & 232 $\pm$ 14 & 311 $\pm$ 8 & 0.92 $\pm$ 0.01 & 1.01 $\pm$ 0.01 & 1694.8 \\
pim & euclidean & 0.21 $\pm$ 0.0 & 266 $\pm$ 41 & 445 $\pm$ 45 & 0.11 $\pm$ 0.0 & 3.59 $\pm$ 0.03 & 6.8 \\
 & monge & 0.29 $\pm$ 0.01 & 515 $\pm$ 72 & 4656 $\pm$ 269 & 2.38 $\pm$ 0.03 & 0.53 $\pm$ 0.02 & 1711.7 \\
 & fisher & 0.21 $\pm$ 0.0 & 427 $\pm$ 47 & 547 $\pm$ 30 & 0.93 $\pm$ 0.0 & 0.98 $\pm$ 0.01 & 425.4 \\
rip & euclidean & 0.09 $\pm$ 0.01 & 829 $\pm$ 112 & 1499 $\pm$ 56 & 0.24 $\pm$ 0.01 & 2.46 $\pm$ 0.02 & 4.2 \\
 & monge & 0.15 $\pm$ 0.02 & 1042 $\pm$ 211 & 3123 $\pm$ 685 & 1.38 $\pm$ 0.02 & 0.97 $\pm$ 0.01 & 593.9 \\
 & fisher & 0.09 $\pm$ 0.0 & 1623 $\pm$ 113 & 1753 $\pm$ 92 & 0.91 $\pm$ 0.0 & 0.96 $\pm$ 0.01 & 204.7 \\
\bottomrule
\end{tabular}
    \caption{Bayesian Logistic Regression. Entries are reported as $\mathrm{mean} \pm \mathrm{std}$.}
\label{tab:logreg}
\end{table}

\section{Mathematical  Derivations} \label{app:math}

\subsection{The Generative and Inverse Generative metrics}

The Generative and Inverse Generative metrics read
\begin{equation*}
    G(x) = f(x) I = \exp (\log f(x) )I,
\end{equation*}
where the scalar factor is $f(x) = \left( \frac{p_0 + \lambda}{p(x) + \lambda} \right)^2$ for the Generative metric and  $f(x) = \left( \frac{p(x) + \lambda}{p_0 + \lambda} \right)^2$ for the Inverse Generative metric.

\paragraph{Square root and inverse square root}
The quantities are given by:
\begin{align*}
        G^{\tfrac{1}{2}}(x) &= \exp{\tfrac{1}{2} \log f(x)} I,\\
        G^{-\tfrac{1}{2}}(x) &= \exp{-\tfrac{1}{2} \log f(x)} I, \\
        \log |\det G(\bx)|&=  D \log f(x).
\end{align*}

\paragraph{Christoffel symbols derivation}
Given the Riemannian metric $G(x) = f(x) I$, the tensor entries are:
\begin{equation*}
G_{ij}(x) = f(x) \delta_{ij}.    
\end{equation*}

The Christoffel symbols for this metric are given by:
\begin{align*}
\Gamma^k_{ij} &= \frac{1}{2 f(x)} \left( \delta_{jk} \partial_i f(x) + \delta_{ik} \partial_j f(x) - \delta_{ij} \partial_k f(x) \right) \\
&= \frac{1}{2} \left( \delta_{jk} \partial_i \log f(x) + \delta_{ik} \partial_j \log f(x) - \delta_{ij} \partial_k \log f(x) \right).
\end{align*}
Denote by $e_k$ the standard basis vectors, the Christoffel symbols in matrix notation are: 
\begin{equation*}
\Gamma^k = \frac{1}{2} \left( \nabla \log f(x) \, e_k^\top + e_k \, \nabla \log f(x)^\top - \partial_k \log f(x) I \right).
\end{equation*}
We compute $\norm{\bbv}^2_{\Gamma^k}=\sum_{i,j=1}^D \Gamma^k_{ij}(\bx) \bbv_i \bbv_j$ which appears in the geodesic equations,
\begin{equation*}
    \norm{\bbv}^2_{\Gamma^k} = \inp{\bbv}{\nabla\log f} \bbv_k - \tfrac{1}{2} \norm{\bbv}^2 \partial_k \log f.
\end{equation*}
The geodesic equations read:
\begin{align*}
    \dot \bx &= \bbv, \\
    \dot \bbv &= \tfrac{1}{2} \norm{\bbv}^2 \nabla \log f-\inp{\bbv}{\nabla\log f} \bbv.
\end{align*}

% \paragraph{ Computation  $\nabla \log f$}
% First let us consider the Inverse Generative metric, recall $f= \left( \frac{p(x) + \lambda}{p_0 + \lambda} \right)^2$.
% The partial derivative of $f(x)$   with respect to $x_i$ is:
% \begin{equation*}
% \partial_i f(x) = \frac{d}{dx_i} \left( \frac{p(x) + \lambda}{p_0 + \lambda} \right)^2 = 2 \left( \frac{p(x) + \lambda}{p_0 + \lambda} \right)    
% \end{equation*}

\subsection{The Monge and Inverse Monge metrics}

The Monge metric and the Inverse Monge metric are:
\begin{equation*}    
    G(x) = I + \alpha^2 \nabla \ell\nabla \ell^\top,  \quad
    G^{-1}(x) = 
    I - \frac{\alpha^2}{1+\alpha^2\norm{\nabla \ell}^2} \nabla \ell\nabla \ell^\top.
\end{equation*}

\paragraph{Square root and inverse square root}
Define the quantity $L_\alpha := 1+\alpha^2 \norm{\nabla\ell}^2$, we list the quantities derived from the matrix and present later their derivation,
\begin{align*}
    G^{1/2}(x) &= I + 
    \frac{\alpha^2}{1+\sqrt{L_\alpha}}
   \nabla\ell(\bx)\nabla\ell(\bx)^\top, \\
    G^{-1/2}(x) &= 
    I-\frac{\alpha^2}{L_\alpha+  \sqrt{L_\alpha}} \nabla\ell(x) \nabla\ell(x)^\top, \\
    \log |\det G(\bx)| &= \log (1+\alpha^2 \norm{\nabla\ell}^2).
\end{align*}
We now derive the quantities.
\citet{Hartmann2022}  gave
\begin{equation}
 G^{-\tfrac{1}{2}}(x) = I +\frac{1}{\norm{\nabla\ell}^2}\left(
\frac{1}{\sqrt{L_\alpha}}-1\right)\nabla\ell\nabla\ell^\top   
\label{eq:monge_1/2_hartmann}
\end{equation}
Note that  if $\norm{\nabla\ell}^2\to 0$ then Equation~\ref{eq:monge_1/2_hartmann} is undefined.
We find a more numerical stable form of $G^{-\tfrac{1}{2}}(x)$ 
Multiply the scalar $\frac{1}{\norm{\nabla\ell}^2}\left(
\frac{1}{\sqrt{L_\alpha}}-1\right)$ by its conjugate
\begin{align*}
  \frac{1}{\norm{\nabla\ell}^2}\left(
\frac{1}{\sqrt{L_\alpha}} -1
\right)   = \frac{1}{\norm{\nabla\ell}^2}\left(
\frac{1-\sqrt{L_\alpha}}{\sqrt{L_\alpha}}\right)
\left(\frac{1+ \sqrt{L_\alpha}}{1+ \sqrt{L_\alpha}} 
\right) = \frac{-\alpha^2}{  L_\alpha + \sqrt{L_\alpha}}.
\end{align*}
Plugging the scalar $\frac{-\alpha^2}{  L_\alpha + \sqrt{L_\alpha}}$ into $G^{-\tfrac{1}{2}}(\bx)$, then it is numerically stable for $\norm{\nabla\ell}^2\to 0$. \\ 
\paragraph{The computation of $G^{\tfrac{1}{2}}(x)$} 
For convenience take $y = \nabla\ell(x)$. The the metric is 
$G(y) = I + y y^\top$. Let us assume the square root is of the form
 $G^{\tfrac{1}{2}}(y) = I + \lambda y y^\top$. Let us formulate a quadratic equation for $\lambda$: 
\begin{align*}
    G^{\tfrac{1}{2}}(y) G^{\tfrac{1}{2}}(y) &=  I + y y^\top\\
    I + 2 \lambda yy^\top + \lambda^2 \norm{y}^2yy^\top &= I + y y^\top \\
    0 &= \left(
        1-2 \lambda - \lambda^2 \norm{y}^2
    \right)  y y^\top.
\end{align*}
The solutions of the quadratic equation are
\begin{equation*}
\lambda = \frac{-1 \pm \sqrt{1 + \|y\|^2}}{\|y\|^2}.
\end{equation*}
Let us simplify $\frac{-1 + \sqrt{1 + \|y\|^2}}{\|y\|^2}$, multiply by its conjugate
\begin{align*}
    \frac{  \sqrt{1 + \|y\|^2}-1}{\|y\|^2}
    \left(\frac{  \sqrt{1 + \|y\|^2}+1}{\sqrt{1 + \|y\|^2}+1}\right)  = 
    \frac{\norm{y}^2}{\norm{y}^2\sqrt{1 + \|y\|^2} + 1} .
\end{align*}
Substitute $y = \alpha\nabla\ell(\bx)$ and we obtain the result
\begin{equation*}
    G^{1/2}(x)= I + \frac{\alpha^2}{\sqrt{1 + \alpha^2\|\nabla\ell(\bx)\|^2} + 1} \nabla\ell\nabla\ell^\top.
\end{equation*}

\paragraph{Christoffel symbols of the Monge metric}
The Christoffel associated to the Monge metric (derivation in Section~\ref{sec:mon} and \citet{Hartmann2022}) are
\begin{align*}
    \Gamma^k(x) &= \frac{\alpha^2}{1+\alpha^2\norm{\nabla\ell}^2}\nabla^2\ell \partial_k \ell,
\end{align*}
and the geodesic equations read
\begin{align*}
    \dot \bx &= \bbv, \\
    \dot \bbv &= -\frac{\alpha^2}{L_\alpha}\norm{\bbv}^2_{\nabla^2\ell} \nabla\ell.
\end{align*}

\paragraph{Christoffel symbols of the Inverse Monge metric} The Christoffel symbols associated to the inverse Monge metric (derivation in Section~\ref{sec:invmon}) are
\begin{equation*}
    \Gamma^k = \frac{\alpha^2}{2} \Big[
L_\alpha \left( \nabla f \nabla \ell^\top + \nabla \ell \nabla f^\top + 2f \nabla^2 \ell \right) \partial_k \ell
+  \nabla \ell \nabla \ell^\top \partial_k f
+ \alpha^2 \inp{\nabla \ell}{\nabla f} \nabla \ell \nabla \ell^\top \partial_k \ell 
\Big],
\end{equation*}
and the geodesic equations read
\begin{align*}
    \dot \bx &= \bbv, \\
    \dot \bbv &= - \frac{\alpha^2}{2} \Big[
2L_\alpha \left(         
        \inp{v}{\nabla f} \inp{\nabla\ell}{v} +         
        f\norm{v}^2_{\nabla^2\ell}
    \right) 
        \nabla \ell
+ \inp{\nabla\ell}{v}^2 \nabla f
+ \alpha^2 \inp{\nabla \ell}{\nabla f} \inp{\nabla\ell}{v}^2
\nabla \ell 
\Big].
\end{align*}

\subsection{Monge metric: Christoffel symbols derivation} \label{sec:mon}
For completeness let us do an alternative derivation of the Christoffel symbols from the one found in \citet{Hartmann2022}. 
Take the auxiliary function $f(x) = -\frac{1}{L_\alpha}$, where $L_\alpha = 1 + \alpha^2 \norm{\nabla\ell}^2$. The metric and inverse components are:
\begin{align*}
    g_{ij} &= \delta_{ij} + \alpha^2 \partial_i \ell\partial_j \ell,\\
    g^{km} &= \delta_{km} + \alpha^2 f(x) \partial_k \ell\partial_m \ell.
\end{align*}
The derivatives of the metric are:
\begin{align*}
    \partial_i g_{mj} &= \alpha^2\left( 
    \partial_{im} \ell\partial_j \ell +  \partial_{m} \ell\partial_{ij} \ell \right),\\
    \partial_j g_{im} &= \alpha^2\left( 
     \partial_{ij} \ell\partial_m \ell +  \partial_{i} \ell\partial_{jm} \ell     \right),\\ 
    \partial_m g_{ij} &= \alpha^2\left( 
     \partial_{im} \ell\partial_j \ell + \partial_{i} \ell\partial_{jm} \ell     \right).
\end{align*}
The Christoffel symbols read, 
\begin{align*}
    \Gamma^k_{ij} &= \frac{1}{2}g^{km}\left(
        \partial_i g_{mj} + 
        \partial_j g_{im} -
        \partial_m g_{ij}
    \right) \\
    &= \frac{\alpha^2}{2}g^{km}\left(             
             2  \partial_{m} \ell\partial_{ij} \ell
    \right) \\
    &= \alpha^2 \sum_m \left(
        \delta_{km} + \alpha^2 f(x)  \partial_k \ell\partial_m \ell
    \right)        
               \partial_{m} \ell\partial_{ij} \ell
     \\
    & = \alpha^2 \left(
        \partial_{k} \ell\partial_{ij} \ell
         + \alpha^2 f(x) \sum_m (\partial_m \ell)^2 \partial_k \partial_{ij} \ell
    \right)\\
    &=  \alpha^2 \partial_{k} \ell\partial_{ij}  \left(
    1 -  \tfrac{\alpha^2 \norm{\nabla\ell}^2}{1+\alpha^2 \norm{\nabla\ell}^2}
        \right)\\
    &= \frac{\alpha^2}{1+\alpha^2 \norm{\nabla\ell}^2} \partial_{k} \ell\partial_{ij}.
\end{align*}
Thus, the Christoffel symbols are $\Gamma^k_{ij} = \frac{\alpha^2}{L_\alpha} \partial_{k} \ell\partial_{ij}$.
Writing in matrix form $\Gamma^k$ of size $D\times D$ with components $[\Gamma^k]_{ij} = \Gamma^k_{ij}$,
\begin{equation*}
\Gamma^k = \frac{\alpha^2}{1+\alpha^2 \norm{\nabla\ell}^2} \nabla^2\ell \partial_{k} \ell.
\end{equation*}
Let us compute $\norm{\bbv}^2_{\Gamma^k}$, which appears in the geodesic equations,
\begin{align*}
    v^\top \Gamma^k v &= 
    \frac{\alpha^2}{1+\alpha^2 \norm{\nabla\ell}^2} \norm{v}^2_{\nabla^2\ell }\partial_{k}\ell. 
\end{align*}
The geodesic equations read:
\begin{align*}
    \dot \bx &= \bbv, \\
    \dot \bbv &= - \frac{\alpha^2}{L_\alpha} \norm{\bbv}^2_{\nabla^2\ell }\nabla \ell.  
\end{align*}

\subsection{Inverse Monge metric: Christoffel symbols derivation} \label{sec:invmon}
Again the auxilary function is $f(x) = -\frac{1}{L_\alpha}$, where $L_\alpha = 1 + \alpha^2 \norm{\nabla\ell}^2$, the metric and inverse components are
\begin{align*}
    g_{ij} &= \delta_{ij} + f(x)\alpha^2 \partial_i \ell\partial_j \ell\\
    g^{km} &= \delta_{km} + \alpha^2 \partial_k \ell\partial_m \ell.
\end{align*}
The derivatives of the metric are (we mark with the same color repeating terms)
\begin{align*}
    \partial_i g_{mj} &= \alpha^2\left(\partial_i f \partial_m \ell\partial_j \ell + 
    \textcolor{blue}{f \partial_{im} \ell\partial_j \ell} + \color{magenta}{f \partial_{m} \ell\partial_{ij} \ell} \right),\\
    \partial_j g_{im} &= \alpha^2\left(\partial_j f \partial_i \ell\partial_m \ell + 
    \color{magenta}{f \partial_{ij} \ell\partial_m \ell} + \textcolor{red}{f \partial_{i} \ell\partial_{jm} \ell}     \right),\\ 
    \partial_m g_{ij} &= \alpha^2\left(\partial_m f \partial_i \ell\partial_j \ell + 
    \textcolor{blue}{f \partial_{mi} \ell\partial_j \ell} + \textcolor{red}{f \partial_{i} \ell\partial_{mj} \ell  }   \right).
\end{align*}
Let us compute the Christoffel symbols of the first kind (blue and red terms will cancel out, pink terms add to each other)
\begin{align*}
   \Gamma_{kij} &= \frac{1}{2}\left( \partial_i g_{mj} + 
        \partial_j g_{im} -
        \partial_m g_{ij} \right) \\
        &=
        \frac{\alpha^2}{2}\left(\partial_i f \partial_m \ell\partial_j \ell + 
            \partial_j f \partial_i \ell\partial_m \ell - 
             \partial_m f \partial_i \ell\partial_j \ell + 
             \textcolor{magenta}{2 f \partial_{ij} \ell\partial_{m} \ell}
             \right).
\end{align*}
The Christoffel symbols of the second kind read
\begin{align*}
    \Gamma^k_{ij} &= \frac{1}{2}g^{km}\left(
        \partial_i g_{mj} + 
        \partial_j g_{im} -
        \partial_m g_{ij}
    \right) \\
    &= \frac{\alpha^2}{2} \sum_m \left(
        \delta_{km} + \alpha^2 \partial_k \ell\partial_m \ell
    \right)
    \left(
            \partial_i f \partial_m \ell\partial_j \ell + 
            \partial_j f \partial_i \ell\partial_m \ell -
             \partial_m f \partial_i \ell\partial_j \ell + 
             2 f  \partial_{ij} \ell \partial_{m} \ell
    \right) \\    
    & = \frac{\alpha^2}{2} \bigg(
        \partial_i f \partial_k \ell\partial_j \ell + 
        \partial_j f \partial_i \ell\partial_k \ell -
         \partial_k f \partial_i \ell\partial_j \ell + 
         2 f  \partial_{ij} \ell \partial_{k} \ell\\
         & \quad+ \alpha^2 \partial_k \ell 
            \left(
                \partial_i f\norm{\nabla\ell}^2\partial_j \ell + 
                \partial_j f \partial_i\ell \norm{\nabla\ell}^2  -
                \inp{\nabla f}{ \nabla \ell} \partial_i \ell\partial_j \ell + 
                2 f  \partial_{ij} \ell\norm{\nabla\ell}^2 
            \right)
    \bigg)\\
    & = \frac{\alpha^2}{2}   \bigg( 
            \partial_k \ell \left(
        L_\alpha  \partial_i f \partial_j \ell
        + L_\alpha \partial_i \ell \partial_j f
        -  \alpha^2 \inp{\nabla f}{ \nabla \ell}
        \partial_i \ell\partial_j \ell
        + 2  L_\alpha f(x) \partial_{ij} \ell 
        \right) - \partial_k f\partial_i \ell\partial_j \ell
    \bigg).   \\
\end{align*}
Thus, the Christoffel symbols are:
\begin{equation*}
  \Gamma^k_{ij} = \frac{\alpha^2}{2}  \Big[
    \partial_k \ell \left(
L_\alpha 
    \left(  
           \partial_i f \partial_j \ell
           +\partial_i \ell \partial_j f
           +2f(x) \partial_{ij} \ell 
    \right) 
 - 
  \alpha^2 \inp{\nabla f}{ \nabla \ell}\partial_i \ell\partial_j \ell
    \right)
    - \partial_k f\partial_i \ell\partial_j \ell
\Big].  
\end{equation*}
Written in matrix form
\begin{equation*}
  \Gamma^k = \frac{\alpha^2}{2}
      \Big[
      \partial_k \ell  \left(
        L_\alpha 
        \left(  
               \nabla f \nabla \ell^\top
               +\nabla \ell \nabla f^\top
               +2f(x) \nabla^2 \ell 
        \right) 
     - 
      \alpha^2 \inp{\nabla f}{ \nabla \ell}\nabla \ell \nabla\ell ^\top
      \right)
      - \partial_k f\nabla \ell \nabla\ell ^\top
    \Big].
\end{equation*}
Let us compute $\norm{v}^2_{\Gamma^k}$ which appears in the geodesic equations
\begin{align*}
    v^\top \Gamma^k v &= 
    \frac{\alpha^2}{2}   \Big[
    \partial_k\ell\left(
2L_\alpha \left(         
        \inp{v}{\nabla f} \inp{\nabla\ell}{v} +         
        f\norm{v}^2_{\nabla^2\ell}
    \right) 
        -  \alpha^2 \inp{\nabla f}{ \nabla \ell}
        \inp{\nabla\ell}{v}^2
        \right)
    - \partial_k f \inp{\nabla\ell}{v}^2
\Big].
\end{align*}
Then the geodesic equations read,
\begin{align*}
    \dot \bx &= \bbv, \\
    \dot \bbv &= - \frac{\alpha^2}{2} \Big[
    \left(
2L_\alpha \left(         
        \inp{v}{\nabla f} \inp{\nabla\ell}{v} +         
        f\norm{v}^2_{\nabla^2\ell}
    \right) 
        - \alpha^2 \inp{\nabla f}{ \nabla \ell}
        \inp{\nabla\ell}{v}^2
        \right)\nabla \ell 
         -\inp{\nabla\ell}{v}^2\nabla f
\Big].
\end{align*}
Where the gradient of $f$ is:
\begin{equation*}
    \nabla f = \frac{2\alpha^2}{L^2_\alpha}\nabla^2\ell \nabla\ell.
\end{equation*}

\subsection{Target Distributions} \label{app:toydist}
The Funnel, Squiggle and Rosenbrock distributions are smooth bijective transformations from a $Z\sim \cN(\mu, \Sigma)$ to $X = f(X)$. We use the shorthand notation $x = x(z) $ and $z =z(x)$.
\paragraph{The Funnel distribution}
$p(x)  = \mathcal{N}(x_D|0,\sigma^2) \mathcal{N}(x_{1:{D-1}}|\mu, e^{x_D}I_{D-1} )$.  In this case $Z\sim \cN(0,I)$. The choice of parameters is $\sigma=3$ and $\mu=0$,
\begin{equation*}
    x = \begin{bmatrix}
    e^{\sigma z_D /2} z_{1:D-1}\\
    \sigma z_D
    \end{bmatrix},
    \quad
    \pdv{x}{z} = \begin{bmatrix}
    e^{\sigma z_D /2} I_{D-1} & \frac{\sigma}{2} e^{\sigma z_D /2} z_{1:D-1} \\
    0 & \sigma
\end{bmatrix}    
\quad 
    \pdv{z}{x} = 
    \begin{bmatrix}
        e^{- x_D /2} I_{D-1}  & -\tfrac{1}{2}e^{- x_D /2}x_{1:D-1} \\
        0 &  \tfrac{1}{\sigma}.
    \end{bmatrix}.
\end{equation*}
The log determinant of the inverse Jacobian is $\log \det(\pdv{z}{x}) =  -(D-1)x_D/2 - \log\sigma$.

\paragraph{The hybrid Rosenbrock distribution}
For simplicity here we show the two dimensional case, the full distribution can be consulted in \citet{Pagani2022}.
The two dimensional density is: $ p(x) = \cN(x_{1}|a, \frac{1}{2})
\cN(x_{2}|x_{1}^2,\frac{1}{2b})$.  In this case $Z\sim \cN(0,I)$. The choice of parameters is $a=1$, $b=100$ and block size of $3$ and $\lfloor \tfrac{D-1}{3}\rfloor$ total blocks,
\begin{equation*}
    x = 
    \begin{bmatrix}
    a + \tfrac{1}{\sqrt{2}}z_1 \\
     (a + \tfrac{1}{\sqrt{2}}z_1)^2 + \tfrac{1}{\sqrt{2b}}z_2
    \end{bmatrix},
    \quad
    \pdv{x}{z} = 
    \begin{bmatrix}
    \frac{1}{\sqrt{2}} & 0 \\
      \sqrt{2} a + z_1 & \frac{1}{\sqrt{2b}}
    \end{bmatrix},
    \quad
    \pdv{z}{x} = 
    \begin{bmatrix}
    \sqrt{2} & 0 \\
    -2\sqrt{2b}x_1 & \sqrt{2b}
    \end{bmatrix}.
\end{equation*}

\paragraph{The Squiggle distribution}
The density is $p(x) = \cN(x(z)|\mu, \Sigma)|\det \pdv{x}{z}|$, where $Z\sim \cN(\mu,\Sigma)$. 
The choice of parameters is $a = 1.5$, $\mu=0$, $\Sigma = \diag(5,\tfrac{1}{2},..,\tfrac{1}{2})$
\begin{equation*}
    x = 
    \begin{bmatrix}
    z_1 \\
     z_{2:D} -\sin(a z_1)
     \end{bmatrix} \quad 
     \pdv{x}{z} =
    \begin{bmatrix}
    1 & 0 \\
      -a \cos(a z_1) & I
    \end{bmatrix},
    \quad
    \pdv{z}{x} = 
    \begin{bmatrix}
    1 & 0 \\
    a \cos(a x_1) & I
    \end{bmatrix}.
\end{equation*}
The log determinant of the inverse Jacobian is $\log \det(\pdv{z}{x}) = 0$.

For these three toy problems the Fisher Information follows from the transformation rule of Riemannian metrics
\begin{equation*}
    G(x) = \pdv{z}{x}^\top  \Sigma^{-1} \pdv{z}{x}.    
\end{equation*}

\paragraph{Location and Scale parameters for Complex Distributions}
In Experiment~\ref{sec:exp3} we consider the mixture of two complex distributions. We introduce a location and scale parameter for each components of the mixture. 
The Funnel, Squiggle and Rosenbrock distributions are smooth bijective transformations from a $Z\sim \cN(\mu, \Sigma)$ to $Y = f(X)$.
Let us add a location and scale parameters by an additional transformation $g(Y) = X$, where $g(y) = \Sigma_{y} y + \mu_{y}$
\begin{equation*} 
     Z \overset{f}{\xmapsto{\hspace{0.6cm}}} Y \overset{g}{\xmapsto{\hspace{0.6cm}}} X.
\end{equation*}
The change of variable formula for the composition $g\circ f$ gives 
\begin{align*}
    p_X(x) &= p_Z \left( (g\circ f)^{-1}(x) \right) \left| \det\pdv{x}{z}\right| \\
        &=  p_Z \left( (f^{-1} \circ g^{-1})(x) \right) \left| \det \pdv{x}{y}\right|\left| \det\pdv{y}{z}\right|
\end{align*}
Plug in $g^{-1}(x) = \Sigma_y^{-1/2} (x-\mu_y) $, $\det \pdv{y}{z} = \det \Sigma^{-1/2}$, and $p_Z(z) = \cN(z|\mu, \Sigma)$, we obtain the expression of the density 
% with a location and scale parameters $(\mu_y, \Sigma_y)$:
\begin{equation*}
    p_X(x) = \cN\left( f^{-1} (\Sigma^{1/2} (x-\mu_y)) \bigg\vert \mu, \Sigma \right) \left| \det \pdv{x}{y}\right|\left| \det \Sigma_y^{-1/2} \right|.
\end{equation*}
Where $(\mu_y, \Sigma_y)$ are the location and scale parameters of the component of the mixture distribution.
% \paragraph{The  Allen-Cahn Field System}
% We consider the stochastic Allen–Cahn model \citep{Berglund2017} used as a benchmark in  \citet{Cabezas2024}. The log-density is:
% \begin{equation}
%     \log p(x) = -\beta \left( \frac{a}{2\Delta s} \sum_{i=1}^{D+1} (x_i - x_{i-1})^2 + \frac{b \Delta s}{4} \sum_{i=1}^{D} (1 - x_i^2)^2 \right). \label{eq:fieldsystem}
% \end{equation}
%  Similar to previous work we consider parameters $\Delta s = \tfrac{1}{D}$, and boundary conditions $x_0=x_{D=1}=0$. The values $a=0.1$ and $b=\tfrac{1}{a}$ are chosen to ensure bimodality for each $x_i$. We set the dimension $D=16$. 
 
%  An inspection of the density of the model (Eq.~\eqref{eq:fieldsystem}) reveals that the global maxima are $(1,..,1)$ and $(-1,..,-1)$, at these values the first terms in the sum  cancel out and the second term is zero. 
%  Note that the second term induces bimodality for $x_i = \pm 1$. Then combinations of values of the form
%  $(\pm 1,..,\pm 1)$ are local maxima, since the second term is zero but the first terms of the sum do not cancel out. Thus the problem has a total of $2^D$ maxima. 

\paragraph{The Allen-Cahn Field System}
We consider the stochastic Allen–Cahn model \citep{Berglund2017} used as a benchmark in \citet{Cabezas2024}. The log-density is:
\begin{equation}
    \log p(x) = -\beta \left( \frac{a}{2\Delta s} \sum_{i=1}^{D+1} (x_i - x_{i-1})^2 + \frac{b \Delta s}{4} \sum_{i=1}^{D} (1 - x_i^2)^2 \right). \label{eq:fieldsystem}
\end{equation}
We adopt the  parameter choices  $\Delta s = \tfrac{1}{D}$ and boundary conditions $x_0 = x_{D+1} = 0$, and the constants $a=0.1$ and $b=\tfrac{1}{a}$ ensure that the double-well potential induces bimodality in each component $x_i$, and we fix $D = 16$.

\paragraph{Analysis of multimodality}
To understand the maxima of this density, we analyze the two terms in the log-density function (Eq.~\ref{eq:fieldsystem}):
\begin{enumerate}
    \item The first term, $ 
    \sum_{i=1}^{D+1} (x_i - x_{i-1})^2$, penalizes differences between adjacent components, encouraging all components to have similar values.
    \item The second term, $\sum_{i=1}^{D} (1 - x_i^2)^2$, is minimized when $x_i = \pm 1$.
\end{enumerate}

The global maxima occur at $(1,\ldots,1)$ and $(-1,\ldots,-1)$ because these configurations minimize both terms simultaneously: all components have the same value (satisfying the first term) and each component equals $\pm 1$ (satisfying the second term).
    
Local maxima occur at all other combinations of $\pm 1$ values (i.e., at points $(\pm 1,\ldots,\pm 1)$ with mixed signs) because these configurations still satisfy the second term perfectly, but incur penalties from the first term due to sign changes between adjacent components.

This creates $2^D$ local maxima, making the problem highly multimodal, with the two homogeneous configurations being global maxima.

\paragraph{Kernel Stein Discrepancy}
Let $\pi$ and $\nu$ be two probability measures. We estimate the Kernel Stein Discrepancy with the biased but non-negative V-estimator. Given a sample $x_i\sim \nu$ for $i=1,\ldots,n$,
\begin{equation*}
    \widehat{\mathrm{KSD}}^2_{k, V}(\pi, \nu) = \frac{1}{n^2} \sum_{i=1}^n \sum_{j=1}^n k_\pi(x_i, x_j').
\end{equation*}
Denote by $p(x)$ be the density of the measure $\pi$, then,
\begin{equation*}
k_\pi(x, x') = \nabla_x \cdot \nabla_{x'} k(x, x') + \nabla_x k(x, x') \cdot \nabla_{x'} \log p(x') + \nabla_{x'} k(x, x') \cdot \nabla_x \log p(x) + k(x, x') \nabla_x \log p(x) \cdot \nabla_{x'} \log p(x),
\end{equation*}
where we choose the inverse multi quadratic kernel $k(x,x') = (1+(x-x')^\top(x-x'))^\beta$ for $\beta=-\tfrac{1}{2}$, following the choices made by \citet{Cabezas2024}.

\end{document}